\documentclass{article}
\usepackage{iclr2027_conference,times}
\iclrfinalcopy

\usepackage{amsmath,amsfonts,bm}

\def\eqref#1{equation~\ref{#1}}

\def\1{\bm{1}}

\DeclareMathAlphabet{\mathsfit}{\encodingdefault}{\sfdefault}{m}{sl}
\SetMathAlphabet{\mathsfit}{bold}{\encodingdefault}{\sfdefault}{bx}{n}

\usepackage{hyperref}
\usepackage{url}
\usepackage{amsmath}
\usepackage{amssymb}
\usepackage{amsthm}
\usepackage{graphicx}
\usepackage{xcolor}
\usepackage{float}
\usepackage{placeins}
\usepackage{booktabs}
\usepackage{multirow}
\usepackage{algorithm}
\usepackage{algpseudocode}
\usepackage{tikz}
\usepackage{helvet}
\usepackage{siunitx}
\usepackage{colortbl}
\definecolor{oursrow}{gray}{0.92}
\usetikzlibrary{positioning, calc, arrows.meta, shapes.geometric, patterns}

\newcommand{\y}{\mathbf{y}}
\newcommand{\z}{\mathbf{z}}
\newcommand{\x}{\mathbf{x}}
\newcommand{\qh}{\hat q}
\newcommand{\GAR}{GAR}
\newcommand{\GARfull}{Gated Answer Resampling}
\newcommand{\Tblk}{T_{\mathrm{blk}}}
\newcommand{\Nsup}{N_{\mathrm{sup}}}
\newcommand{\yinit}{\y_{\mathrm{init}}}
\newcommand{\zinit}{\z_{\mathrm{init}}}

\newcommand{\fitwidth}[1]{\resizebox{\ifdim\width>\linewidth\linewidth\else\width\fi}{!}{#1}}
\title{Resample the Answer: Where to Spend\\ Test-Time Compute in Recursive Reasoning}

\author{Maxime Illouz \\
Concordia University \\
Mila -- Qu\'ebec AI Institute \\
Montr\'eal, Canada \\
\texttt{maxime.illouz@mila.quebec}
}

\begin{document}

\maketitle
\lhead{Preprint}

\begin{abstract}
Recursive reasoning models such as the Tiny Recursive Model (TRM) iteratively refine two latent states: an answer state $\y$, decoded into the output, and a reasoning state $\z$, which is never decoded. Many of their failures have been attributed to bad basins, regions of latent space that trap the recursion on wrong answers. Recent test-time approaches perturb $\y$, $\z$ or both to recover these failures, but the impact of this choice on performance, test-time compute efficiency and scaling properties remains unclear. Through controlled experiments on TRM, we find that a failed trajectory carries its failure in its answer state. At equal compute, whether retries continue a failed trajectory (depth scaling) or draw multiple trajectories from the initial states (width scaling), resampling $\y$ recovers more failures than any perturbation of $\z$ we tested, and the performance gap widens as the test-time compute budget grows. Our approach, \textbf{Gated Answer Resampling (GAR)}, pairs this resampling of the answer state with the model's own halting head and restarts only the puzzles that the head rejects. With at most $64$ restarts per puzzle and no training or hyperparameter search, GAR takes TRM from $85.9\%$ to $99.4\%$ on Sudoku-Extreme and from $68.8\%$ to $85.6\%$ on Pencil Puzzle Bench validation. On the same checkpoints, GAR also outperforms Probabilistic TRM (PTRM), which perturbs $\z$ at each step, even when PTRM uses $42$ to $197$ times as much compute. Our code is available at \url{https://github.com/MaximeILL/GAR}.
\end{abstract}

\section{Introduction}
\label{sec:intro}

Recursive reasoning models such as the Hierarchical Reasoning Model \citep[HRM;][]{wang2025hrm} and the
Tiny Recursive Model \citep[TRM;][]{jolicoeur2025trm} solve hard structured puzzles by refining a latent state over many
iterations instead of writing out intermediate steps as tokens. TRM does so with one two-layer network of a few
million parameters and two states: an \emph{answer state} $\y$, decoded into the output grid, and a
\emph{reasoning state} $\z$, which is never decoded. In each latent recursion, $\z$ is
recomputed six times from the input, the answer state and itself, after which $\y$ is updated once from $\y$ and $\z$. This makes $\y$, the state the decoder reads, the slow state and the only one the input never reaches directly. Since the recursion is deterministic, a rerun fails the same
puzzles in exactly the same way. \citet{sghaier2026ptrm} attribute many of these failures to
\emph{bad basins}, regions of latent space that decode to wrong grids and that the recursion does
not escape, and \citet{ren2026mechanistic} report a similar failure mode in HRM, spurious fixed points
of its latent dynamics that correspond to wrong answers.

Recent test-time approaches make TRM's trajectories stochastic. Probabilistic TRM (PTRM) adds Gaussian noise
to $\z$ before every supervision step of $K$ independent rollouts so that they follow different trajectories and some reach a good basin \citep{sghaier2026ptrm}.
\citet{efstathiou2026attractor} perturb the initial answer as a proof of concept, and \citet{corbett2026guided} run a
particle filter whose proposal perturbs both states. All three rely on the model's halting head to
select or reweight trajectories. GRAM (Generative Recursive reAsoning Models; \citealp{baek2026gram}) and Equilibrium Reasoners (EqR; \citealp{huang2026eqr}), by contrast, train new models with noise, on the slow state for GRAM and on both states for EqR. Three questions remain open. First, these approaches perturb $\y$, $\z$ or both, but it remains unclear which state carries the failure, and what impact this choice has on performance, test-time efficiency and scaling properties. In particular, it is unknown whether test-time compute is better spent diversifying trajectories through $\z$, as PTRM does, or resampling $\y$. Evidence from training is mixed: noise on the fast state did not help GRAM (its footnote~3), randomising either initial state alone gave TRM similar gains in EqR's ablation (single runs), and \citet{dillon2026speed} finds that training works best with four different reasoning states per puzzle and one shared answer state. Second, it is also unclear whether extra test-time compute should continue a failed trajectory (depth) or restart it from the initial state (width). PTRM, for instance, scales depth without noise and width with noise, and its longer noisy rollouts start from the initial state rather than from a failed trajectory. Third, all five draw fresh
trajectories for every puzzle, even those the model already solves, and PTRM's headline
configurations cost $400$ and $300$ standard passes.

\definecolor{navy}{RGB}{47,74,148}
\definecolor{blk}{RGB}{237,241,250}
\definecolor{zc}{RGB}{52,103,201}
\definecolor{yc}{RGB}{224,113,37}
\definecolor{gc}{RGB}{40,140,96}
\definecolor{gr}{RGB}{128,132,140}
\definecolor{ink}{RGB}{40,44,52}
\def\tk{0.19}
\newcommand{\ztokens}[3]{%
  \foreach \i in {0,...,5}{
    \ifnum#3=0
      \draw[draw=zc!75!black, fill=zc!40, line width=0.35pt, rounded corners=0.6pt] (#1+\i*\tk,#2) rectangle ++(\tk-0.03,\tk-0.03);
    \else
      \draw[draw=zc!80, fill=white, line width=0.5pt, rounded corners=0.6pt] (#1+\i*\tk,#2) rectangle ++(\tk-0.03,\tk-0.03);
    \fi}}
\newcommand{\ytokens}[3]{%
  \foreach \i in {0,...,5}{
    \pgfmathparse{int(mod(\i,3))}
    \ifnum\pgfmathresult=0
      \ifnum#3=0
        \draw[draw=gr!85!black, fill=gr!35, line width=0.35pt, rounded corners=0.6pt] (#1+\i*\tk,#2) rectangle ++(\tk-0.03,\tk-0.03);
      \else
        \draw[draw=yc!80, fill=white, line width=0.5pt, rounded corners=0.6pt] (#1+\i*\tk,#2) rectangle ++(\tk-0.03,\tk-0.03);
      \fi
    \else
      \ifnum#3=0
        \draw[draw=yc!75!black, fill=yc!45, line width=0.35pt, rounded corners=0.6pt] (#1+\i*\tk,#2) rectangle ++(\tk-0.03,\tk-0.03);
      \else
        \draw[draw=yc!80!black, fill=white, dashed, dash pattern=on 1pt off 0.7pt, line width=0.45pt, rounded corners=0.6pt] (#1+\i*\tk,#2) rectangle ++(\tk-0.03,\tk-0.03);
      \fi
    \fi}}

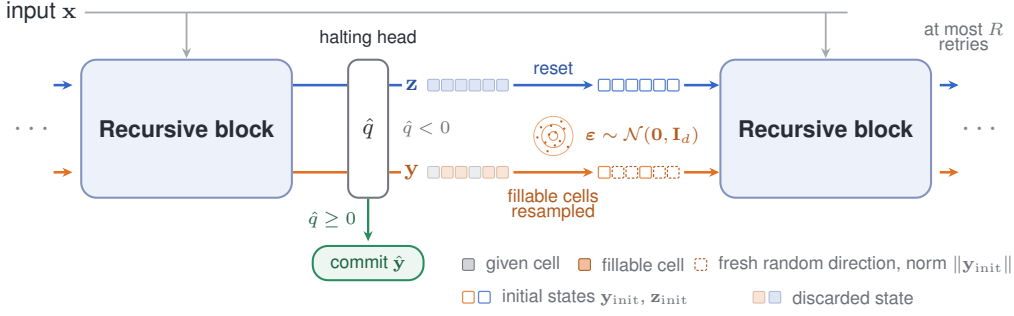
\begin{figure}[t]
\centering
\resizebox{0.97\textwidth}{!}{\sffamily\selectfont%
\begin{tikzpicture}[
  x=1cm,y=1cm, font=\sffamily\scriptsize, text=gr!85!black,
  >={Stealth[length=4pt,width=3.6pt]},
  block/.style={draw=navy!60, fill=blk, rounded corners=6pt, line width=0.8pt, minimum width=2.9cm, minimum height=1.9cm, align=center},
  gate/.style={draw=gr!90!black, fill=white, rounded corners=4pt, line width=0.8pt, minimum width=0.55cm, minimum height=1.9cm, align=center, text=ink, font=\footnotesize},
  commit/.style={rounded corners=6pt, draw=gc, fill=gc!8, line width=0.8pt, inner xsep=7pt, inner ysep=3.5pt, text=gc!70!black},
  lane/.style={line width=1.0pt},
]
\def\zy{0.60}  \def\yy{-0.60}
\def\bA{2.55}  \def\gx{5.05}  \def\bB{11.35}
\node[text=ink, font=\footnotesize] at (0.55,1.6) {input $\mathbf{x}$};
\draw[draw=gr!70, line width=0.7pt] (1.15,1.6) -- (\bB+0.3,1.6);
\draw[draw=gr!70, line width=0.7pt, ->] (\bA,1.6) -- (\bA,0.97);
\draw[draw=gr!70, line width=0.7pt, ->] (\bB,1.6) -- (\bB,0.97);
\node[block] (B1) at (\bA,0) {\textbf{\footnotesize\textcolor{ink}{Recursive block}}};
\node[block] (B2) at (\bB,0) {\textbf{\footnotesize\textcolor{ink}{Recursive block}}};
\node[text=gr, font=\large] at (\bA-2.1,0) {$\cdots$};
\node[text=gr, font=\large] at (\bB+2.15,0) {$\cdots$};
\draw[lane, draw=zc, ->, shorten >=3pt] (\bA-1.83,\zy) -- (B1.west |- 0,\zy);
\draw[lane, draw=yc, ->, shorten >=3pt] (\bA-1.83,\yy) -- (B1.west |- 0,\yy);
\draw[lane, draw=zc, ->, shorten <=3pt] (B2.east |- 0,\zy) -- (\bB+1.83,\zy);
\draw[lane, draw=yc, ->, shorten <=3pt] (B2.east |- 0,\yy) -- (\bB+1.83,\yy);
\node[gate] (G) at (\gx,0) {$\hat q$};
\node[anchor=south, text=ink] at (\gx,1.0) {halting head};
\draw[lane, draw=zc] (B1.east |- 0,\zy) -- (G.west |- 0,\zy);
\draw[lane, draw=yc] (B1.east |- 0,\yy) -- (G.west |- 0,\yy);
\draw[draw=gc, line width=0.9pt, ->] (G.south) -- (\gx,-1.55) node[midway, left=1pt, text=gc!70!black] {$\hat q \ge 0$};
\node[commit, anchor=north] at (\gx,-1.6) {commit $\hat{\mathbf{y}}$};
\draw[lane, draw=zc] (G.east |- 0,\zy) -- (5.58,\zy);
\node[anchor=center, text=zc!80!black, font=\footnotesize, fill=white, inner sep=1.2pt] at (5.65,\zy) {$\mathbf{z}$};
\begin{scope}[opacity=0.4]\ztokens{5.87}{\zy-0.08}{0}\end{scope}
\draw[lane, draw=zc, ->] (7.05,\zy) -- (8.15,\zy) node[midway, above=0.5pt, text=zc!80!black] {reset};
\ztokens{8.22}{\zy-0.08}{1}
\draw[lane, draw=zc, ->] (9.40,\zy) -- (B2.west |- 0,\zy);

\draw[lane, draw=yc] (G.east |- 0,\yy) -- (5.58,\yy);
\node[anchor=center, text=yc!80!black, font=\footnotesize, fill=white, inner sep=1.2pt] at (5.65,\yy) {$\mathbf{y}$};
\begin{scope}[opacity=0.4]\ytokens{5.87}{\yy-0.08}{0}\end{scope}
\draw[lane, draw=yc, ->] (7.05,\yy) -- (8.15,\yy) node[midway, below=2.5pt, align=center, text=yc!80!black] {fillable cells\\[-2pt] resampled};
\ytokens{8.22}{\yy-0.08}{1}
\draw[lane, draw=yc, ->] (9.40,\yy) -- (B2.west |- 0,\yy);
\begin{scope}[shift={(7.60,-0.10)}]
  \draw[draw=yc!60, line width=0.35pt] (0,0) circle (0.27);
  \draw[draw=yc!75, line width=0.35pt] (0,0) circle (0.16);
  \draw[draw=yc!85, line width=0.35pt] (0,0) circle (0.06);
  \foreach \p in {(0.03,0.05),(-0.06,0.02),(0.09,-0.07),(-0.11,-0.09),(0.02,-0.13),(-0.03,0.14),(0.15,0.10),(-0.17,0.12),(0.20,-0.15),(-0.08,-0.21),(0.11,0.20),(-0.22,-0.02)}{ \fill[yc!85!black] \p circle (0.016); }
\end{scope}
\node[anchor=west, text=yc!80!black] at (7.92,-0.10) {$\boldsymbol\varepsilon \sim \mathcal N(\mathbf 0,\mathbf I_d)$};
\node[anchor=west, text=gr!85!black] at ($(G.east)+(0.06,0)$) {$\hat q<0$};
\node[anchor=south, align=center, text=gr!85!black] at (\bB+1.9,0.95) {at most $R$\\[-2pt] retries};
\draw[draw=gr!85!black, fill=gr!35, line width=0.35pt, rounded corners=0.6pt] (6.35,-1.95) rectangle ++(\tk-0.03,\tk-0.03); \node[anchor=west] at (6.55,-1.87) {given cell};
\draw[draw=yc!75!black, fill=yc!45, line width=0.35pt, rounded corners=0.6pt] (7.95,-1.95) rectangle ++(\tk-0.03,\tk-0.03); \node[anchor=west] at (8.15,-1.87) {fillable cell};
\draw[draw=yc!80!black, fill=white, dashed, dash pattern=on 1pt off 0.7pt, line width=0.45pt, rounded corners=0.6pt] (9.55,-1.95) rectangle ++(\tk-0.03,\tk-0.03); \node[anchor=west] at (9.75,-1.87) {fresh random direction, norm $\lVert\mathbf{y}_{\mathrm{init}}\rVert$};
\draw[draw=yc!80, fill=white, line width=0.5pt, rounded corners=0.6pt] (6.35,-2.40) rectangle ++(\tk-0.03,\tk-0.03);
\draw[draw=zc!80, fill=white, line width=0.5pt, rounded corners=0.6pt] (6.57,-2.40) rectangle ++(\tk-0.03,\tk-0.03); \node[anchor=west] at (6.77,-2.32) {initial states $\mathbf{y}_{\mathrm{init}}$, $\mathbf{z}_{\mathrm{init}}$};
\begin{scope}[opacity=0.4]\draw[draw=yc!75!black, fill=yc!45, line width=0.35pt, rounded corners=0.6pt] (10.35,-2.40) rectangle ++(\tk-0.03,\tk-0.03); \draw[draw=zc!75!black, fill=zc!40, line width=0.35pt, rounded corners=0.6pt] (10.57,-2.40) rectangle ++(\tk-0.03,\tk-0.03);\end{scope} \node[anchor=west] at (10.77,-2.32) {discarded state};
\end{tikzpicture}%
}
\caption{\textbf{\GARfull{} (\GAR{}).} The recursion runs in blocks of $16$ supervision steps over a fast reasoning state $\mathbf{z}$ (blue) and a slow answer state $\mathbf{y}$ (orange). The first block is the standard pass. After each block, the grid $\hat{\mathbf{y}}$ is committed if the halting logit $\hat{q}$, read from $\mathbf{y}$, is non-negative. Otherwise a new block (a \emph{retry}) runs, at most $R$ times. A retry restarts from the initial states, except on the fillable cells of $\mathbf{y}$, which receive fresh isotropic Gaussian vectors rescaled to the norm of $\mathbf{y}_{\mathrm{init}}$. If no grid is committed, the \emph{fallback} returns the grid of the block with the highest $\hat{q}$.}
\label{fig:grz}
\end{figure}

We study these questions within a shared \emph{harness}: after the standard pass, each puzzle that the halting head rejects receives up to $R$ \emph{retries}, extra runs of $16$ supervision steps, until the head accepts a grid. Only the retry varies: it perturbs $\y$, $\z$ or both, and it continues the failed trajectory or restarts from the initial state. Our study leads to \textbf{\GARfull{} (\GAR)}, a test-time compute approach whose retries restart from the initial state, with the fillable cells of $\y$ (those not fixed by the input) resampled as isotropic Gaussian vectors rescaled to the norm of the initial answer state (Figure~\ref{fig:grz} and Algorithm~\ref{alg:gated}).
We make three contributions.

\begin{enumerate}
\itemsep0.3em
\item \textbf{The answer state carries the failure.} After $16$ retries, continuing the failed trajectory without perturbation recovers $74\%$ of the standard-pass failures on Sudoku-Extreme and $16\%$ on Pencil Puzzle Bench (PPBench) validation. Resampling $\y$ raises this to $89\%$ and $34\%$, whereas resampling $\z$ barely helps ($74\%$ and $19\%$). A diagnostic suggests why, qualifying TRM's view that $\z$ acts like a chain of thought: one step into a continued retry, resampling $\y$ has moved $\z$ further than resampling $\z$ itself, as if $\z$ were largely recomputed from the input and $\y$. Once $\y$ is resampled, restarting from the initial state performs as well ($88\%$ and $40\%$), so retries can restart and run in parallel (Figure~\ref{fig:grid}, Section~\ref{sec:where}).

\item \textbf{For both depth and width scaling, test-time compute is better spent resampling $\y$ than perturbing $\z$, increasingly so as the budget grows.} At the same number of retries, whether they continue the failed trajectory (depth scaling) or draw multiple trajectories from the initial states (width scaling), resampling $\y$ once per retry recovers more failures than any perturbation of $\z$ we tested, even PTRM's noise before every step, and the gap widens as the compute budget increases (Figures~\ref{fig:grid} and~\ref{fig:attempts}). After $64$ restarts, it recovers $96\%$ of the standard-pass failures on Sudoku-Extreme and $54\%$ on PPBench validation, compared with $85\%$ and $27\%$ when $\z$ is resampled instead, and $85\%$ and $33\%$ for PTRM itself, on the same checkpoints and under the shared harness. With only $16$ restarts, it already recovers $88\%$ and $40\%$, more than either perturbation of $\z$ does with $64$ (Table~\ref{tab:iso}).

\item \textbf{The resulting approach, \GAR{}, is tuning-free and compute-efficient.} On Sudoku-Extreme and PPBench validation, it restarts only the puzzles the halting head rejects: $99.3\%$ and $87.5\%$ of the failures and none of the successes, so it loses no solved puzzle (Table~\ref{tab:gate}). With up to $64$ restarts and no noise scale to tune, it raises TRM from $85.9\%$ to $99.4\%$ and from $68.8\%$ to $85.6\%$, at only $2.03\times$ and $7.08\times$ the cost of a standard pass. On the same checkpoints, PTRM stays below \GAR{} even with $197$ and $42$ times as much compute under its own protocol, and $4.3$ and $6.0$ times as much under the shared harness (Section~\ref{sec:gar}).
\end{enumerate}

\section{Background}
\label{sec:background}

\paragraph{Tiny Recursive Models.} TRM \citep{jolicoeur2025trm} keeps two latent states of shape
$[P + L, d]$, with $L$ grid cells preceded by a prefix of $P$ positions ($P = 16$ in the released
implementation) and hidden width $d = 512$. The answer state $\y$ is decoded into the grid $\hat\y$
by a linear head on its cell positions, and the reasoning state $\z$ is never decoded. A
\emph{supervision step} applies one two-layer network $f_\theta$ in $T = 3$ latent recursions,
each made of $n = 6$ updates of $\z$ and then one update of $\y$,
\begin{equation}
\label{eq:recursion}
\z \leftarrow f_\theta(\x + \y + \z) \quad (n \text{ times}), \qquad \y \leftarrow f_\theta(\y + \z),
\end{equation}
where $\x$ is the input embedding. Under \emph{deep supervision}, the detached $(\y, \z)$ initialise
the next step, for at most $\Nsup = 16$ steps, with a cross-entropy loss on the output logits at
every step. A second linear head, the \emph{halting head}, reads the first prefix position of $\y$
and outputs the halting logit $\qh$, trained at every step by binary cross-entropy against the
exact-match indicator. This simplified form of HRM's adaptive computation time
\citep[ACT;][]{wang2025hrm, graves2016act} halts a training sample once $\qh > 0$, and is switched off at
evaluation, where all $\Nsup$ steps run \citep{sghaier2026ptrm}. In the released implementation,
each layer of $f_\theta$ ends in an RMS normalisation without learnable gain, so every position of
$\y$ and $\z$ that $f_\theta$ outputs has norm $\sqrt d$. The initial states are fixed, untrained
random vectors broadcast to every position, and they are not normalised. We write $\yinit, \zinit
\in \mathbb{R}^d$ for these vectors and for their broadcast to all $P + L$ positions, and take norms
of states per position. One run of $\Tblk = \Nsup = 16$ steps from a given state is a
\emph{block}. The standard evaluation, which we call the \emph{standard pass}, is one block from
$(\yinit, \zinit)$, and all costs below are in blocks per puzzle.

\paragraph{Depth and width.} \citet{sghaier2026ptrm} distinguish two axes of test-time compute.
\emph{Depth} continues one trajectory beyond the $\Nsup$ steps the model was trained on, and
\emph{width} draws several trajectories per puzzle from the initial state. Depth can be scaled
without perturbation. Doing so recovers most failures on Sudoku-Extreme but few on PPBench (Section~\ref{sec:held}), where PTRM also reports only a small gain.
Width cannot be scaled at all without perturbation, since a deterministic model replays its trajectory bit for bit. To scale width, PTRM runs $K$ rollouts per puzzle from $(\yinit, \zinit)$, adds $\boldsymbol\epsilon \sim \mathcal{N}(\mathbf
0, \sigma^2 \mathbf I)$ to $\z$ before every supervision step and, after $D$ steps, returns the grid
of the rollout with the largest $\qh$ (\emph{best-$Q$@$K$}, \emph{pass@$K$} is the oracle), for
$KD/16$ blocks per puzzle. With $D > 16$, it also scales depth. Its headline configurations on
Sudoku-Extreme ($K = 100$, $D = 64$) and PPBench ($K = 100$, $D = 48$) cost $400$ and $300$ blocks.

\section{A Controlled Space of Retries}
\label{sec:protocol}

\paragraph{Shared harness.} Under the harness, every method runs the standard pass from $(\yinit,
\zinit)$ and reads the halting logit at its last supervision step. If $\qh \ge 0$, the grid is committed and
nothing else runs. Otherwise, the \emph{gate} re-opens the puzzle, and retries run one block at a
time, for at most $R$ blocks. The first retry with $\qh \ge 0$ ends the search (the \emph{stopping
rule}), and its grid is committed. If none reaches $\qh \ge 0$, the harness returns the grid with the highest $\qh$ among the standard pass and the retries. This \emph{fallback}
keeps the standard-pass grid unless a retry is rated above it. We fixed two budgets in advance, the default $R = \Nsup = 16$ and $R = 64$. Since the random draws of a retry do not depend on $R$, the first $r$
retries of any run form a run with $R = r$, which is what \emph{after $r$ retries} means. A solved
puzzle can be lost only if the gate re-opens it (a false positive). On Sudoku-Extreme, the gate re-opens $99.3\%$ of the puzzles the standard pass fails and none of those it solves. On PPBench, it re-opens $87.5\%$ of the failures and none of the successes on validation, and $93.8\%$ and $3.0\%$ on the golden split (Table~\ref{tab:gate}).

\paragraph{Retries.} A retry either \emph{continues} the failed trajectory from where the
previous block left it, which scales depth, or \emph{restarts} from $(\yinit, \zinit)$, which scales
width. It either applies no perturbation, \emph{resamples} one state or, as in PTRM, adds noise to $\z$
before every supervision step. We also run retries that perturb both states, but only when continuing (Section~\ref{sec:carry}). To resample a state, we replace the vector of each of its fillable cells, once at the block boundary, with a fresh direction, drawn from an isotropic Gaussian and rescaled to the norm of that state's initial value,
$\lVert\yinit\rVert$ or $\lVert\zinit\rVert$. On Sudoku-Extreme, the fillable cells, $\mathcal{M}(\x)$,
are the blank cells. PPBench has no blank token, so Appendix~\ref{app:protocol} defines its fillable cells. The other cells, fixed by the input, are the given cells. Positions a retry does not resample
are kept if it continues and set back to $(\yinit, \zinit)$ if it restarts. Restarts are thus independent draws, so they can also run in parallel. Keeping the first accepted grid in the order of the draws, or the fallback's grid if none is accepted, then gives the same grids as the sequential run. Running them one at a time only lets the stopping rule save compute (Appendix~\ref{app:gatedwidth}). \GAR{} restarts and resamples $\y$, and the two central comparisons each change one
thing in it: they resample $\z$ instead of $\y$, or continue instead of restarting (the
\emph{continuation variant} of \GAR). In Algorithm~\ref{alg:gated}, skipping line~5 gives the continuation variant, and skipping lines~5 and~6 gives the continuation without perturbation, which is the depth baseline. Skipping line~6 alone gives a restart without perturbation, which replays the standard pass bit for bit (a control). Resampling $\z$ instead of $\y$ in line~6 gives the restart that resamples $\z$. Skipping line~6 and adding noise to $\z$ before every step of the \textsc{block} in line~7 gives PTRM under the harness. To compare methods at matched numbers of retries, we run it with one rollout of $D = 16$ steps per retry. The restart that resamples $\z$ then lets us split the gap between \GAR{} and PTRM, puzzle by puzzle, into two paired differences: which state is resampled, and whether $\z$ is resampled once or receives noise before every step (Section~\ref{sec:form}).

\begin{algorithm}[t]
\caption{\textbf{\GARfull{} (\GAR{}).} \textsc{block} runs $\Tblk$ supervision steps and returns the new state, the grid and the last halting logit, without reading any label.}
\label{alg:gated}
\footnotesize
\begin{algorithmic}[1]
\Require puzzle $\x$, a trained TRM, $\Tblk = \Nsup = 16$, budget $R$ ($= \Nsup$ by default), halting threshold $0$
\State $(\y, \z, \hat\y, \qh) \gets \textsc{block}(\x, \yinit, \zinit)$ \Comment{the standard $16$-step pass}
\State $(\hat\y^\star, \qh^\star) \gets (\hat\y, \qh)$ \Comment{fallback on the halting logit}
\For{$r = 1$ \textbf{to} $R$}
  \State \algorithmicif\ $\qh \ge 0$ \algorithmicthen\ \Return $\hat\y^\star$ \Comment{the gate at $r = 1$, then the stopping rule}
  \State $(\y, \z) \gets (\yinit, \zinit)$ \Comment{restart: nothing crosses the boundary}
  \State $\y_i \gets \lVert \yinit \rVert \, \boldsymbol{\varepsilon}_i / \lVert \boldsymbol{\varepsilon}_i \rVert$,
    \ $\boldsymbol{\varepsilon}_i \sim \mathcal{N}(\mathbf{0}, \mathbf{I}_d)$, for every cell $i \in \mathcal{M}(\x)$
    \Comment{prefix and given cells untouched}
  \State $(\y, \z, \hat\y, \qh) \gets \textsc{block}(\x, \y, \z)$
  \State \algorithmicif\ $\qh > \qh^\star$ \algorithmicthen\ $(\hat\y^\star, \qh^\star) \gets (\hat\y, \qh)$
\EndFor
\State \Return $\hat\y^\star$
\end{algorithmic}
\end{algorithm}

\paragraph{Setup.} We train one checkpoint per benchmark (Appendix~\ref{app:checkpoints}): TRM-MLP,
the attention-free variant of TRM \citep{jolicoeur2025trm}, on \emph{Sudoku-Extreme}
\citep{wang2025hrm}, and TRM-Att, the self-attention variant, on six puzzle types of \emph{PPBench}
\citep{waugh2026ppbench}. On Sudoku-Extreme, we evaluate on $1{,}000$ random test puzzles and the
standard pass solves $85.90\%$ ($141$ failures, $140$ re-opened). On PPBench, following PTRM, we
report five of the six types (sudoku, lightup, \mbox{nurikabe}, heyawake and tapa) and the standard
pass solves $68.83\%$ of our $308$ validation puzzles of these five types ($96$ failures, $84$
re-opened). Unless a split is named, PPBench results in Sections~\ref{sec:where} and~\ref{sec:gar}
are on validation, and results on the golden split (the same $49$ puzzles as in PTRM) are in
Appendix~\ref{app:ppbench}.

\paragraph{Baseline.} Since PTRM's checkpoints were not publicly available, all methods, PTRM
included, run on our checkpoints. These differ from PTRM's in standard-pass accuracy, slightly on
Sudoku-Extreme and by several points on PPBench (Appendix~\ref{app:ptrm}), so comparisons across
papers are only indicative. We therefore rerun PTRM's protocol, best-$Q$@$K$ over rollouts drawn for
every puzzle. We draw $K = 100$ rollouts per puzzle and use the first $K' \le K$. We call this rerun
\emph{PTRM with its own protocol}. On Sudoku-Extreme, we keep PTRM's $\sigma = 0.3$. On PPBench, we
use $\sigma^\star = 1.5$, inside the plateau of best-$Q$@$25$ in a sweep of $\sigma$ on our validation
split, where PTRM's value, $0.2$, scores $2.70$ points lower (Table~\ref{tab:sigma-sweep}). The scale
$\sigma^\star$ is thus selected on the labels of the split we report.

\paragraph{Statistical protocol.} Stochastic methods run with three sampling seeds on a fixed checkpoint, and $\pm$ is the standard deviation over seeds. Differences between methods are paired by instance and averaged over seeds, with $95\%$ bootstrap intervals over instances ($B = 10{,}000$). \GAR{} also draws the same random directions as its continuation variant and the restart that resamples $\z$ (Appendix~\ref{app:protocol}). We call a difference significant when its $95\%$ interval excludes zero, without correction for multiplicity.

\section{Where to Explore at Test Time}
\label{sec:where}

\begin{figure}[t]
\centering
\includegraphics[width=\linewidth]{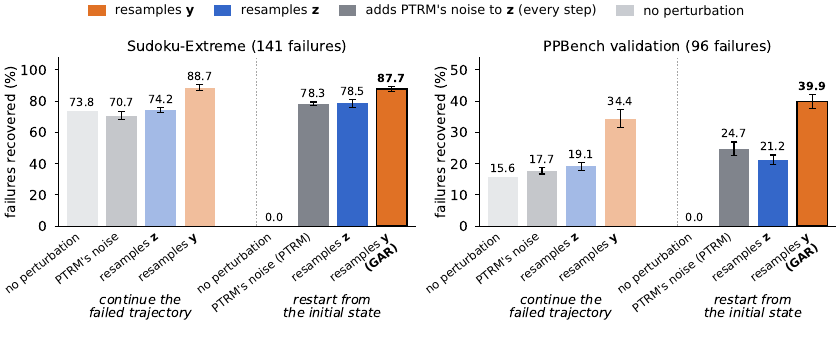}%
\caption{\textbf{Resampling $\y$ recovers the most failures, whether retries continue or restart.} Fraction of the standard-pass failures that end up correct after $16$ retries, out of $141$ on Sudoku-Extreme and $96$ on PPBench validation. In each panel, retries continue the failed trajectory on the left (depth) and restart from the initial state on the right (width). Each side has four bars, one per kind of retry: no perturbation, PTRM's noise on $\z$ (on the right, PTRM itself under the harness), resampling $\z$, and resampling $\y$ (on the right, \GAR{}). A restart without perturbation replays the standard pass and recovers nothing. Bars show the mean over three seeds $\pm$ one standard deviation (values in Table~\ref{tab:grid}).}
\label{fig:grid}
\end{figure}

\paragraph{Depth alone as a baseline.}
\label{sec:held}
Under the harness, after $16$ retries, continuing the recursion beyond the training horizon without perturbation already recovers $73.8\%$ of the standard-pass failures on Sudoku-Extreme, but only $15.6\%$ on PPBench validation (Figure~\ref{fig:grid}). On Sudoku-Extreme, most failures thus come from trajectories that the horizon cuts short. Trajectories that remain trapped in a spurious attractor \citep{efstathiou2026attractor} or cannot escape a bad basin \citep{sghaier2026ptrm} can only be among the failures that depth alone leaves, so they account for at most $26.2\%$ of the failures on Sudoku-Extreme and $84.4\%$ on PPBench. We next ask what keeps the remaining failures in place.

\subsection{Which State to Perturb When Continuing}
\label{sec:inside}

\begin{figure}[h]
\centering
\includegraphics[width=\linewidth]{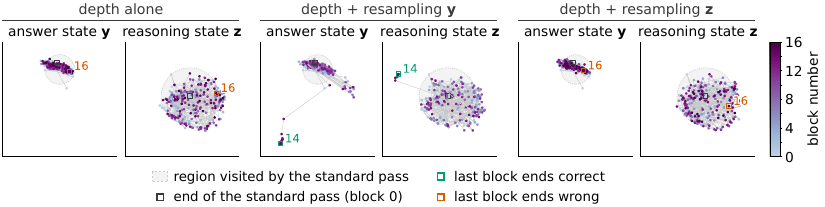}%
\caption{\textbf{When continuing a failed trajectory, resampling $\y$ lets it escape, whereas a resampled $\z$ falls back near its region.} The states $\y$ and $\z$ of one sudoku from the PPBench validation split after each supervision step, coloured by block. Each state is projected onto two principal axes common to the three runs. With depth alone, $\y$ stays near the region visited by the standard pass. Resampled at each block boundary, $\y$ returns there until block $14$, where it escapes, $\z$ follows, and the block ends on the correct grid. Resampled instead, $\z$ is back near its region one step later, and $\y$ stays near its own, as with depth alone.}
\label{fig:escape}
\end{figure}

A continued retry starts from both states of the failed trajectory, so perturbing one of them tests which state carries the failure. Resampling $\y$ removes $56.8 \pm 7.2\%$ of the failures that depth alone leaves on Sudoku-Extreme and $22.2 \pm 3.3\%$ on PPBench (left side of each panel of Figure~\ref{fig:grid}), whereas the same resampling of $\z$ removes $1.8 \pm 5.6\%$ and $4.1 \pm 1.4\%$. Figure~\ref{fig:escape} follows one failure that depth alone leaves and that resampling $\y$ recovers, but resampling $\z$ does not. The gain from resampling $\y$ is positive for every seed on both benchmarks, and it comes at a lower cost, since re-opened puzzles stop sooner (Appendix~\ref{app:lattice-more}). No intervention that we ran on $\z$ alone removes more than $1.8\%$ of the failures that depth alone leaves on Sudoku-Extreme or $5.8\%$ on PPBench, be it PTRM's noise in continued retries (at five scales on PPBench), a reset to $\zinit$ or a resampling of every grid cell (Appendix~\ref{app:lattice-more}). The asymmetry grows with retries. After one retry, resampling $\z$, resampling $\y$ and depth alone have solved $48.0$, $49.7$ and $38$ of the $140$ re-opened Sudoku-Extreme puzzles and after $16$ retries, the counts are $104.7$, $125.0$ and $104$ (Appendix~\ref{app:persistence}). Roughly half of the gain from resampling $\y$ comes from erasing the carried answer. A deterministic reset of $\y$ to $\yinit$ on the fillable cells removes $27.0\%$ and $9.9\%$ of these failures, and the fresh directions account for the rest.

\subsection{Which State to Perturb When Restarting}
\label{sec:restart}

\begin{figure}[t]
\centering
\includegraphics[width=\linewidth]{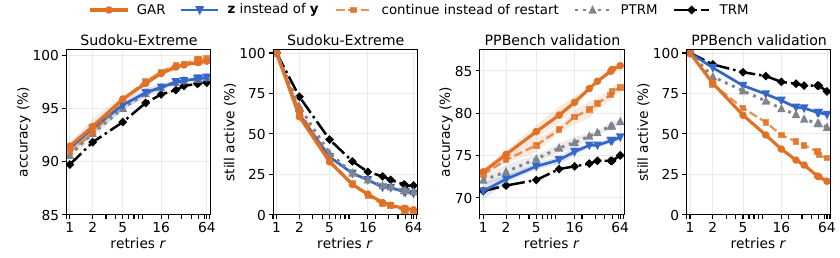}%
\caption{\textbf{Resampling $\y$ scales better with retries than diversifying trajectories through $\z$, at a lower cost.} For each benchmark, the figure shows the accuracy after $r$ retries and the fraction of re-opened puzzles still active (no grid accepted) before the $r$-th retry, against $r$ on a logarithmic scale (Table~\ref{tab:iso-full} gives accuracies and costs at seven of the ten plotted values of $r$). Whether $\z$ is resampled once (``$\z$ instead of $\y$'') or perturbed before every step (PTRM), \GAR{}'s lead grows with $r$, and its searches end sooner. On PPBench, each doubling of $r$ adds about two points to \GAR{}, versus about one when $\z$ is perturbed. On Sudoku-Extreme, ``continue instead of restart'' almost coincides with \GAR{}. Curves show the mean over three seeds $\pm$ one standard deviation. TRM continues without perturbation under the harness (depth alone). Like \GAR{} and ``$\z$ instead of $\y$'', PTRM scales width: each retry is one rollout of $D = 16$ steps.}
\label{fig:attempts}
\end{figure}

A restart discards both states of the failed trajectory, so it tests the choice of state when no failure is carried over. Resampling the answer state still leads (Figures~\ref{fig:grid} and~\ref{fig:attempts}): after $64$ retries, \GAR{} recovers $96\%$ of the standard-pass failures on Sudoku-Extreme and $54\%$ on PPBench, versus $85\%$ and $27\%$ for the same restart with $\z$ resampled instead of $\y$ (Table~\ref{tab:iso}). The gap is significant at $16$ and $64$ retries on both benchmarks (Table~\ref{tab:iso-terms}), and it widens as the budget increases. Resampling $\z$ also costs more (Table~\ref{tab:iso}) and leaves more re-opened puzzles still active before the $64$th retry: $13.1\%$ vs.~$2.9\%$ for \GAR{} on Sudoku-Extreme, and $61.9\%$ vs.~$20.6\%$ on PPBench (Figure~\ref{fig:attempts}). At $64$ retries, \GAR{} solves nearly every puzzle that this restart solves: summed over the three seeds, it solves $49$ of the $52$ puzzles that only one of the two solves on Sudoku-Extreme, and $84$ of $90$ on PPBench (Appendix~\ref{app:iso}).

\paragraph{Resampling once or adding noise at every step.}
\label{sec:form}
When a restart perturbs $\z$, resampling it once and adding PTRM's noise before every step both recover $85\%$ of the Sudoku-Extreme failures after $64$ retries. On PPBench, with a noise scale selected on the labels of the split we report, PTRM recovers $33\%$ vs.~$27\%$ for a single resampling, a significant difference, but only about a fifth of the gap between resampling $\y$ and resampling $\z$ (Tables~\ref{tab:iso} and~\ref{tab:iso-terms}). Which state is perturbed thus matters far more than how $\z$ is perturbed. The two perturbations of $\z$ also differ in size and extent: the resampling rotates the fillable cells by $90^\circ$, whereas the noise deflects every position, prefix included, by $16.7^\circ$ on Sudoku-Extreme and $56.3^\circ$ on PPBench before each step (Appendices~\ref{app:protocol} and~\ref{app:ptrm}). Their comparison therefore does not isolate how often $\z$ is perturbed.

\begin{table}[t]
\caption{\textbf{With $16$ retries, \GAR{} already outperforms TRM and both perturbations of $\z$ at $64$ retries.} Accuracies (\%, $\pm$ standard deviation) and costs in blocks per puzzle (in parentheses) are means over three seeds. All methods, PTRM included, share our checkpoints, instances and harness, which begins with the standard pass of TRM, given in the header. Under the harness, TRM continues without perturbation (depth alone). Each indented row changes one thing in \GAR{}. PTRM runs one rollout of $D = 16$ steps per retry and differs from ``$\z$ instead of $\y$'' only in how $\z$ is perturbed.}
\label{tab:iso}
\centering
\small
\setlength{\tabcolsep}{3pt}
\fitwidth{%
\begin{tabular}{@{}l cc c cc@{}}
\toprule
& \multicolumn{2}{c}{\emph{Sudoku-Extreme}} & & \multicolumn{2}{c}{\emph{PPBench validation}} \\
& \multicolumn{2}{c}{TRM, standard pass: $85.90$} & & \multicolumn{2}{c}{TRM, standard pass: $68.83$} \\
\cmidrule{2-3} \cmidrule{5-6}
& $16$ retries & $64$ retries & & $16$ retries & $64$ retries \\
\midrule
TRM, under the harness & $96.30$\phantom{{\scriptsize$\,\pm0.00$}}~$(1.97)$ & $97.40$\phantom{{\scriptsize$\,\pm0.00$}}~$(3.33)$ &  & $73.70$\phantom{{\scriptsize$\,\pm0.00$}}~$(4.79)$ & $75.00$\phantom{{\scriptsize$\,\pm0.00$}}~$(15.20)$ \\
PTRM & $96.93${\scriptsize$\,\pm0.15$}~$(1.82)$ & $97.87${\scriptsize$\,\pm0.06$}~$(2.90)$ &  & $76.52${\scriptsize$\,\pm0.68$}~$(4.25)$ & $79.00${\scriptsize$\,\pm0.19$}~$(11.86)$ \\
\midrule
\rowcolor{oursrow}\textbf{\GAR} & $98.27${\scriptsize$\,\pm0.21$}~$(1.68)$ & $99.43${\scriptsize$\,\pm0.15$}~$(2.03)$ &  & $81.28${\scriptsize$\,\pm0.68$}~$(3.48)$ & $85.61${\scriptsize$\,\pm0.37$}~\phantom{0}$(7.08)$ \\
\quad $\z$ instead of $\y$ & $96.97${\scriptsize$\,\pm0.35$}~$(1.80)$ & $97.90${\scriptsize$\,\pm0.00$}~$(2.84)$ &  & $75.43${\scriptsize$\,\pm0.50$}~$(4.42)$ & $77.16${\scriptsize$\,\pm0.19$}~$(12.92)$ \\
\quad continue instead of restart & $98.40${\scriptsize$\,\pm0.26$}~$(1.69)$ & $99.73${\scriptsize$\,\pm0.06$}~$(1.98)$ &  & $79.55${\scriptsize$\,\pm0.86$}~$(3.73)$ & $83.01${\scriptsize$\,\pm0.50$}~\phantom{0}$(9.06)$ \\
\bottomrule
\end{tabular}}
\end{table}

\subsection{Continuing or Restarting}
\label{sec:carry}

With $\y$ resampled, continuing differs from restarting only in what it carries over from the failed block: all of $\z$, and the prefix and given cells of $\y$. Table~\ref{tab:lattice-full} cross-tabulates what a continued retry does to $\z$ and to $\y$, and suggests how little $\z$ carries across a block boundary. Once $\y$ is resampled, also perturbing $\z$ (by a resampling, a reset to $\zinit$ or PTRM's noise) recovers $87\%$ to $90\%$ of the standard-pass failures on Sudoku-Extreme after $16$ retries and $31\%$ to $33\%$ on PPBench, compared with $89\%$ and $34\%$ for resampling $\y$ alone. On PPBench, even returning $\z$ to $\zinit$ on every position ($32\%$) stays below \GAR{} ($40\%$), which also resets the prefix and given cells of $\y$. With the same random directions, however, continuing instead of restarting makes no significant difference at $16$ retries on either benchmark (Table~\ref{tab:iso-terms}). At $64$ retries, continuing takes recovery from $96\%$ to $98\%$ on Sudoku-Extreme, a barely significant gain, and from $54\%$ to $45\%$ on PPBench, a significant loss, nearly all on its sudoku type (Appendix~\ref{app:ppbench}). Retries can thus restart and, being independent, run in parallel. Run one at a time, restarts are also cheaper on PPBench (Table~\ref{tab:iso}).

\subsection{Why the Answer State?}
\label{sec:why}

In the recursion, $\z$ is the fast state and $\y$ the slow one. Within one supervision step, $\z$ is updated eighteen times, and each of these updates reads $\x$ and $\y$. The answer state is updated only three times (Equation~\ref{eq:recursion}). A resampled $\y$ is therefore read by all seven updates of the next latent recursion, while a resampled $\z$ is read once and then recomputed through the map $\z \mapsto f_\theta(\x + \y + \z)$. \citet[Proposition~5.1]{corbett2026guided} assume that this map contracts locally around the deterministic path, and on both of their models, at moderate noise, the deviations they measure stay within the bound this assumption implies. We hypothesise the same contraction on our checkpoints. Perturbing $\z$ does not change the map, so $\z$ can return towards the unperturbed path. Perturbing $\y$ changes the map itself, and moves $\z$ away from that path. On Sudoku-Extreme, we compare each perturbed continuation with an unperturbed copy run in lockstep. One supervision step after all grid cells of $\z$ are resampled, $\z$ is already back to a cosine of $0.829$ with that copy. At that point, resampling the fillable cells of $\y$ moves $\z$ further than resampling the same cells of $\z$ itself ($0.565$ vs.~$0.837$, Appendix~\ref{app:persistence}). The diagnostic thus suggests that, to a first approximation, whatever $\z$ carries across a block boundary at inference is recomputed from $\x$ and $\y$. This would qualify the view that $\z$ records how the model reached its answer \citep{jolicoeur2025trm}. Inside a failed trajectory, the carried $\y$ would then impose the failure on $\z$ again at every update. From the initial state, where $\yinit$ carries no answer, a resampled $\z$ can still redirect the trajectory even though it is read only once. It recovers fewer failures than a resampled $\y$ (Figure~\ref{fig:grid}), which is read seven times. In the diagnostic, $\z$ is not recomputed exactly. Once $\z$ is resampled, its cosine settles near $0.74$, and by the end of the block the grid has changed about as much as when $\y$ is resampled.

\section{Experimental Evaluation}
\label{sec:gar}

\paragraph{Design and tuning.} \GAR{} (Algorithm~\ref{alg:gated}) runs the harness with retries that restart and resample $\y$, as Section~\ref{sec:where} suggests. It reads no label at inference, and we tune none of its hyperparameters: the block length and the default $R$ come from the training horizon, the threshold from the halting rule, the norm from $\yinit$, and $\mathcal{M}(\x)$ from the input tokens. The only free parameter is the retry budget $R$, which trades compute for accuracy, monotonically on average (Figure~\ref{fig:attempts}). Labels of the reported splits did, however, inform two design decisions: the choice of retry, made on Sudoku-Extreme and PPBench validation, and, on PPBench, the given cells of each puzzle type, which we specified by hand and checked against the labels of its validation and golden splits (Appendix~\ref{app:protocol}).

\begin{table}[ht]
\caption{\textbf{With $R = 64$, \GAR{} outperforms every PTRM configuration we ran and loses no solved puzzle.} We rerun PTRM on our checkpoints, with $\sigma = 0.3$ on Sudoku-Extreme and $\sigma^\star = 1.5$ on PPBench. The first column gives the perturbed state, and \emph{rejected} means the puzzles the halting head rejects. Cost is in blocks per puzzle. \emph{Lost} counts the puzzles that the standard pass solves and a method fails, out of $859$ and $212$. Stochastic rows are averaged over three seeds, and the standard deviation of their accuracy is at most $0.46$ points on Sudoku-Extreme and $0.75$ on PPBench. The uniform unroll runs the recursion for $D$ steps on every puzzle. The dagger marks PTRM's headline budget of $100$ rollouts of $64$ and $48$ steps. It costs $197$ and $42$ times as much as \GAR{} at $R = 64$, and $4.3$ and $6.0$ times as much under the harness. Tables~\ref{tab:sweep} and~\ref{tab:sweep-pp} list ten configurations per benchmark.}
\label{tab:main}
\centering
\small
\setlength{\tabcolsep}{4pt}
\begin{tabular}{@{}c l l S[table-format=3.2] S[table-format=2.2] S[table-format=2.1] c S[table-format=3.2] S[table-format=2.2] S[table-format=2.1]@{}}
\toprule
& & & \multicolumn{3}{c}{\emph{Sudoku-Extreme}} & & \multicolumn{3}{c}{\emph{PPBench validation}} \\
\cmidrule{4-6} \cmidrule{8-10}
\textbf{state} & \textbf{method} & \textbf{runs on} & {\textbf{cost}} & {\textbf{accuracy}} & {\textbf{lost}} & & {\textbf{cost}} & {\textbf{accuracy}} & {\textbf{lost}} \\
\midrule
\multicolumn{10}{@{}l}{\emph{Deterministic TRM}} \\
--- & TRM, standard pass & all & 1 & 85.90 & {---} &  & 1 & 68.83 & {---} \\
--- & TRM, uniform unroll ($D{=}64$/$48$) & all & 4 & 92.80 & 0.0 &  & 3 & 68.18 & 10.0 \\
\midrule
\multicolumn{10}{@{}l}{\emph{PTRM with its own protocol: $K'$ rollouts (width) of $D$ steps (depth)}} \\
$\z$ & PTRM ($K'{=}1$, $D{=}16$) & all & 1 & 87.70 & 29.0 &  & 1 & 68.29 & 12.0 \\
$\z$ & PTRM ($K'{=}1$, $D{=}64$/$48$) & all & 4 & 93.60 & 5.0 &  & 3 & 68.72 & 15.7 \\
$\z$ & PTRM ($K'{=}25$, $D{=}16$) & all & 25 & 97.40 & 1.0 &  & 25 & 77.81 & 1.0 \\
$\z$ & PTRM ($K'{=}100$, $D{=}16$) & all & 100 & 98.07 & 0.3 &  & 100 & 80.63 & 1.3 \\
$\z$ & PTRM ($K'{=}100$, $D{=}64$/$48$)$^\dagger$ & all & 400 & 98.57 & 0.0 &  & 300 & 79.98 & 10.3 \\
\midrule
\multicolumn{10}{@{}l}{\emph{PTRM under the harness: at most $K'$ retries of $D$ steps}} \\
$\z$ & PTRM ($K'{=}100$, $D{=}64$/$48$)$^\dagger$ & rejected & 8.65 & 98.47 & 0.0 &  & 42.25 & 82.14 & 0.0 \\
\midrule
\multicolumn{10}{@{}l}{\emph{\GAR{} (ours): at most $R$ retries of $16$ steps}} \\
\rowcolor{oursrow} $\y$ & \textbf{\GAR{} ($R{=}16$)} & rejected & 1.68 & 98.27 & 0.0 &  & 3.48 & 81.28 & 0.0 \\
\rowcolor{oursrow} $\y$ & \textbf{\GAR{} ($R{=}64$)} & rejected & 2.03 & 99.43 & 0.0 &  & 7.08 & 85.61 & 0.0 \\
\bottomrule
\end{tabular}
\end{table}

\paragraph{With PTRM's own protocol.} Averaged over three seeds, \GAR{} with $R = 16$ is above every PTRM
configuration costing up to $100$ blocks on Sudoku-Extreme, and every configuration on PPBench. With $R = 64$, it even exceeds the pass@$K'$ oracle of every configuration on both benchmarks
(Tables~\ref{tab:sweep} and~\ref{tab:sweep-pp}). This protocol can also lose solved puzzles. For every puzzle, including those the standard pass solves, it returns the grid of the noisy rollout with the largest $\qh$ (best-$Q$@$K'$). With a single noisy rollout of $16$ steps, it loses $3.4\%$ of the solved Sudoku-Extreme puzzles, but with $100$ rollouts almost none. On PPBench, its $300$-block configuration still loses $4.9\%$ of the solved puzzles, about as many as the uniform unroll of $48$ steps (Table~\ref{tab:main}). On Sudoku-Extreme, where the head ranks rollouts almost perfectly, such losses come from the protocol rather than from the head (Appendix~\ref{app:gatedwidth}). \GAR{} loses no solved puzzle on Sudoku-Extreme or PPBench validation, and one on the golden split, re-opened by the gate's only false positive (Tables~\ref{tab:gate} and~\ref{tab:golden}).

\paragraph{Under the harness.} The harness applies to PTRM as well, with its $K'$ rollouts run one at a time, each as one retry of $D$ steps. For PTRM's headline budgets, it divides the cost by $46$ and $7.1$, loses no solved puzzle and changes accuracy by $-0.10$ and $+2.16$ points (Table~\ref{tab:main}). Of the configurations we ran, only these two exceed
\GAR{} at $R = 16$, at $5.1$ and $12$ times its cost, and none reaches \GAR{} at $R = 64$ (Appendix~\ref{app:gatedwidth}). At
matched numbers of $16$-step retries, PTRM stays below \GAR{} on average at every budget and is never cheaper
(Figure~\ref{fig:attempts}, Table~\ref{tab:iso-full}).

\section{Related Work}
\label{sec:related}

\paragraph{Recursive computation and halting.} Iterating a weight-tied module to trade compute for accuracy appears in Universal Transformers \citep{dehghani2018universal}, deep-thinking networks \citep{schwarzschild2021, bansal2022, bear2024} and recurrent-depth language models \citep{geiping2025recurrentdepth}. HRM and TRM add deep supervision and a head that halts over
supervision steps, a segment-level form of ACT \citep{graves2016act}. At inference, HRM raises its
segment budget. GRAM \citep[Appendix~A.1]{baek2026gram} and \citet[Section~6.3]{huang2026eqr} can
keep the head on to stop each trajectory early, and \citet{dillon2026speed} returns the answer of
the first ensemble member to halt. Our harness uses the head differently, to decide which puzzles are re-opened, since its logit stays negative on most failures (\citealp[Section~3.2]{sghaier2026ptrm}; \citealp[Section~5.10]{dillon2026speed}).

\paragraph{Bad basins in recursive reasoners.} Failure modes of iterated updates predate recursive
reasoners: convergence to initialisation-dependent attractors \citep{anil2022pathindependence} and
degradation beyond the trained number of iterations \citep{bansal2022}, which \citet{viakhirev2026shallow} tie to the dynamical regime of the trained operator. For HRM,
\citet{ren2026mechanistic} identify spurious fixed points and scale
the model's guesses with data augmentation, input perturbations at inference and model bootstrapping. For TRM, \citet{sghaier2026ptrm} call the phenomenon
\emph{bad basins}, and \citet{efstathiou2026attractor} read the recursion as a search over an attractor
landscape and attribute failures to the answer initialisation rather than to insufficient depth. Perturbing both latent states, \citet{corbett2026guided}
recover failures with a particle filter and \citet{huang2026eqr} with random restarts after noisy training. \citet{zhao2026erm} select among these restarts with energies built from the task rules. On our
checkpoints, depth alone recovers most Sudoku-Extreme failures (Section~\ref{sec:held}), as \citet{krishna2026finish} find on two released checkpoints, and the initial answer is the stronger lever, not the only one (Section~\ref{sec:restart}).

\paragraph{Where noise is injected.} GRAM learns its stochasticity on the slow state, the counterpart of TRM's $\y$, so its training-time design and our test-time interventions on a model trained without noise pick the same state. \citet{drozdova2026diffusion} train a denoiser whose inference re-noises non-clue variables at every step. \citet[Appendix~A.3]{huang2026eqr} vary which initial state is
randomised in training (single runs: $84.06\%$ with a fixed initialisation, $86.29\%$ and $86.25\%$ with only the answer or only the reasoning state randomised, respectively, and at best $87.30\%$ with both, the reasoning state at a larger scale). \citet[Section~5.4]{dillon2026speed} trains TRMs whose
parallel states compete through a winner-take-all loss, and reports higher accuracy with parallel
reasoning states than with parallel answer states ($97.7\%$ vs.~at most $95.9\%$). Both results are about training, and neither says which state a test-time retry should perturb in a model trained with fixed initial states. PTRM contrasts its noise at every
step with the single perturbation by \citet{efstathiou2026attractor}. Since the first acts on $\z$ and the second on the initial answer, that
contrast varies the state and the form of the perturbation together. The restart that resamples $\z$ isolates the state (Section~\ref{sec:form}).

\section{Conclusion}
\label{sec:conclusion}
 
Our evidence suggests that, in a failed trajectory, it is the slow, decoded answer state that carries the failure. Resampling it removes some of the failures that depth alone leaves, while resampling
the reasoning state instead adds almost nothing. On our TRM checkpoints, test-time compute is better spent on the answer state than on the reasoning state, increasingly so as the budget grows, and only where the model's own halting head rejects its answer. At matched retries and under the shared harness, resampling the answer state beats PTRM itself at a lower cost, and its lead grows with the number of retries. Since restarting does about as well as continuing, width can be scaled with independent retries. \GAR{} applies these findings without training or hyperparameter search. It can lose a solved puzzle only if the gate re-opens it, and with $R = 64$ it outperforms every PTRM configuration we ran, at $2.03$ and $7.08$ standard passes, versus up to $400$ and $300$ for PTRM.
 
\paragraph{Limitations.} Our PPBench checkpoint is trained on a reconstruction of PTRM's data pipeline (Appendix~\ref{app:ppbench-build}). \GAR{} inherits the halting head, which misses $12.5\%$ of the PPBench validation failures and can accept wrong grids. We have not tested \GAR{} on Maze-Hard, where this head is poorly aligned \citep{sghaier2026ptrm, corbett2026guided}, on ARC-AGI \citep{chollet2019measure}, or beyond TRM. On PPBench, its gains are smallest on nurikabe and heyawake (Appendix~\ref{app:ppbench}).

\section*{Reproducibility Statement}
The code, including the PPBench build, is available at \url{https://github.com/MaximeILL/GAR}, and the two checkpoints at \url{https://huggingface.co/MaximeILL/GAR_ckpts}.
Section~\ref{sec:protocol} specifies the protocol, and Appendix~\ref{app:protocol} lists the checks:
bit-identical standard passes across paired runs and a self-test of the restart.
Appendix~\ref{app:ptrm} explains how we rerun PTRM on
our checkpoints, including how its noise scale is chosen. The Sudoku-Extreme checkpoint is trained
with the released recipe of \citet{jolicoeur2025trm}, with AdamW in place of its AdamATan2 optimiser, and our PPBench build is a reconstruction from Appendix~A.3 of
\citet{sghaier2026ptrm}. Appendix~\ref{app:checkpoints} gives the data and the training
configuration of both checkpoints.

\section*{AI Use Statement}
We used generative AI tools as coding assistants for some of our experiments and to edit parts of the paper for readability. We have reviewed all AI-assisted work, and the code runs the checks listed in Appendix~\ref{app:protocol}. We take full responsibility for the final content of this work.

\section*{Acknowledgements} We would like to acknowledge that the author Maxime Illouz is a PhD student funded by bursaries from Canada Research Chair grant number CRC-2023-00228 (\emph{Computing for the Greater Good}) awarded to Elvis Dohmatob.

\bibliography{references}
\bibliographystyle{iclr2027_conference}

\appendix

\section*{Overview of the Appendices}
Appendix~\ref{app:protocol} specifies the protocol and the checks, and
Appendix~\ref{app:checkpoints} describes the data and training of the two checkpoints.
In Appendix~\ref{app:ptrm}, we rerun PTRM on these checkpoints, both with PTRM's own protocol and under the harness. Appendices~\ref{app:lattice-more} to~\ref{app:persistence} contain
the detailed results behind Section~\ref{sec:where}, namely every perturbation we ran, matched retries at
seven budgets, and the lockstep diagnostic. Appendix~\ref{app:ppbench} breaks the PPBench results down by
type and reports the golden split.

\section{Protocol and Checks}
\label{app:protocol}

\paragraph{Evaluation path.} Every method follows the evaluation path of the training code, with batches of $768$, exactly $\Tblk = 16$ steps per
block, and correctness counted by the evaluator of that code. All methods in a run
use the same batches, standard pass and gate. Our implementation advances the whole batch and discards the outputs of committed
rows. Rows do not interact, so up to floating-point rounding this is equivalent to advancing only the rows not yet committed, and cost is charged for those rows only.

\paragraph{Details of the retries.} At each retry, \GAR{} returns every position of $\y$ and $\z$ to
$\yinit$ and $\zinit$, prefix included, just as the model's own reset does at a cold start, and then
resamples the fillable cells of $\y$. No retry resamples the prefix of $\y$, whose first position is the only one the halting head reads. Continued retries leave the prefix and the given cells of $\y$ untouched, and the continuation variant does not touch any position of $\z$. \GAR, its continuation variant and the restart that resamples $\z$ use \emph{common random numbers}: their directions depend only on the sampling seed and the batch, so a puzzle receives the same directions at a given retry in all three (the third applies them to $\z$). The continuation that resamples $\z$ draws its directions from the same distribution, but independently of the other methods.

\paragraph{The mask on PPBench.} Sudoku-Extreme has a blank token, but PPBench has no single token that marks a fillable cell, since each token encodes the whole visible state of a cell. We therefore declare, once per
puzzle type, the \emph{static} clue families, whose answer is fixed by the input. These are the given digits on
sudoku, the walls on lightup and shakashaka, the numbered cells on nurikabe (which stay white) and the clue cells
on tapa (which are never shaded). Heyawake has none, because its numbers belong to rooms and a cell that
shows one can be shaded. The mask $\mathcal{M}(\x)$ is the set of cells whose token is neither the pad token nor in a
static family, and at inference it reads only the input tokens.

\paragraph{Checks.} The standard passes of paired runs match bit for
bit, on grids and logits alike (for the PTRM rollouts on
Sudoku-Extreme, checked on the first batch of $768$ puzzles). Positions output by $f_\theta$ have norm $\sqrt{d} = 22.627$, and resampled directions
are rescaled to $\lVert \yinit \rVert = 21.244$ in $\y$ and $\lVert \zinit \rVert = 22.689$ in $\z$.
Without perturbation, the restart leaves accuracy unchanged on every split, and continued blocks
reproduce a single unrolled recursion on grids and logits (on
Sudoku-Extreme, a $D = 64$ unroll is identical to four continued blocks of $16$). A
self-test on a real batch of each benchmark checks that, when nothing is resampled, the restart replays
the standard pass bit for bit. Geometrically, PTRM's noise at $\sigma = 0.2$, $0.3$ and $1.5$ deflects a position of norm $\sqrt{d}$ by
$11.3^\circ$, $16.7^\circ$ and $56.3^\circ$ and scales its norm by $1.02$, $1.04$ and $1.80$, whereas a resampling at matched norm rotates it by $90^\circ$.

\begin{table}[h]
\caption{\textbf{The halting head at the last step of the standard pass, against ground truth.} Recall is the fraction of the standard-pass failures that the gate re-opens ($140$ of $141$, $84$ of $96$ and $15$ of $16$), and the false-positive rate is the fraction of the standard-pass successes that it re-opens ($0$ of $859$, $0$ of $212$ and $1$ of $33$).}
\label{tab:gate}
\centering
\small
\fitwidth{
\begin{tabular}{llrrrr}
\toprule
benchmark & split & $n$ & standard pass (\%) & recall (\%) & false-positive rate (\%) \\
\midrule
Sudoku-Extreme & $1{,}000$ test puzzles & $1{,}000$ & $85.90$ & $99.3$ & $0.0$ \\
PPBench (5 types) & validation & $308$ & $68.83$ & $87.5$ & $0.0$ \\
PPBench (5 types) & golden & $49$ & $67.35$ & $93.8$ & $3.0$ \\
\bottomrule
\end{tabular}
}
\end{table}

The gate builds on an observation by \citet[Section~3.2]{sghaier2026ptrm}, who find on PPBench that at convergence the halting logit is near $+6$ on correct trajectories and near $-6$ on incorrect
ones.

\FloatBarrier
\section{Data and Checkpoints}
\label{app:checkpoints}

Both checkpoints use the model, loss, data loader and training loop from the released code of
\citet{jolicoeur2025trm}. We leave this code as it is, except for the reporting of the PPBench evaluation
(Appendix~\ref{app:ppbench-build}) and the optimiser of the network weights, where we use AdamW \citep{loshchilov2019adamw}, as named
in the TRM paper, instead of the AdamATan2 optimiser \citep{everett2024scaling} of the released code.
Table~\ref{tab:ckpt-config} gives the training configuration of the two checkpoints. As in the released loader, an epoch
visits every training puzzle once, through one of its examples. Each checkpoint is trained on a single NVIDIA A100 80GB GPU.

\subsection{Sudoku-Extreme}
\label{app:sudoku-build}

We build the data with the released script and command, from the Hugging Face dataset
\texttt{sapientinc/sudoku-extreme}, which is the Sudoku-Extreme release of \citet{wang2025hrm}. The training split holds
$1{,}000$ puzzles, drawn without a fixed seed, each expanded into $1{,}001$ examples (the puzzle itself and
$1{,}000$ copies transformed by rule-preserving symmetries). The test split consists of the $422{,}786$
puzzles of the test file, without augmentation, and the $1{,}000$ puzzles of Section~\ref{sec:protocol} are
drawn from it uniformly without replacement, with seed $0$. We run the released TRM-MLP command (Table~\ref{tab:ckpt-config}), which trains for $50{,}000$ epochs whereas the TRM paper reports $60{,}000$, and keep the last checkpoint. It reaches $85.90\%$ on our $1{,}000$ puzzles and $86.50\%$ on the full test split,
compared with the $87.4\%$ reported by \citet{jolicoeur2025trm}.

\subsection{PPBench}
\label{app:ppbench-build}

\citet{sghaier2026ptrm} describe their PPBench build in prose (their Appendix~A.3) and through per-type
counts (their Table~4), with a unified $294$-token vocabulary over $100$ positions (sudoku being padded to
$10 \times 10$). Neither the tokenisation nor the number of epochs is specified. Our
build is therefore a reconstruction, which may be one reason why our checkpoint differs from theirs.
Table~\ref{tab:ppb-counts} compares our splits with theirs.

\paragraph{Source and splits.} Our corpus is the full PPBench dataset \citep{waugh2026ppbench},
$62{,}231$ puzzles released as the Hugging Face dataset \texttt{bluecoconut/pencil-puzzle-bench}. We
first remove the $90$ golden puzzles of the six types ($15$ per type), matched to their corpus records. As in
PTRM, $100$ puzzles per type ($50$ for tapa) are then drawn for validation before any size filter,
here with seed $0$, and the rest are kept for training. All three splits then keep only $9 \times 9$
sudoku and $10 \times 10$ grids of the other types (Table~\ref{tab:ppb-counts}).

\paragraph{Tokenisation.} Each cell is a single token encoding its full visible state, namely the puzzle type,
the clue (a given digit, a wall with or without a number, a nurikabe or heyawake number, a
tapa clue, or none) and the answer mark (a digit, a shaded cell or a bulb, a shakashaka triangle and
its orientation, or none). Inputs keep the annotation dots of the replayed moves while labels drop
them, so a label is the solved board. A heyawake cell also carries two wall bits for the room borders. Sudoku is padded with a pad token, and the $19$ padded label positions are ignored by the
loss and by every metric. The vocabulary, read off the data, has $439$ ids ($438$ cell states plus the
pad token).

\paragraph{Augmentation.} According to PTRM's text, each training puzzle is expanded into $10$ examples, the
unaugmented pair and $9$ sampled ones. Yet every per-type training count in its Table~4 is a multiple
of $11$ (and four of the six are not multiples of $10$). We therefore use $11$ examples per training puzzle, the unaugmented pair and
$10$ sampled ones. With
this factor, our training and validation puzzles add up exactly to PTRM's for sudoku, lightup, shakashaka and tapa, while nurikabe and heyawake keep $70$ and $257$ more puzzles than PTRM's counts
imply. A sampled example takes as input the board after the first
$k$ moves of the solution ($k$ uniform on $\{0, \dots, m - 1\}$ for $m$ moves), and one of the
$8$ dihedral transforms, drawn uniformly, is applied to both input and label. Validation and golden puzzles are not
augmented.

\paragraph{Training and checkpoint.} We train TRM-Att with the Sudoku-Extreme settings and a puzzle embedding indexed by
puzzle type, for $21{,}000$ epochs (Table~\ref{tab:ckpt-config}). A puzzle counts as solved when every supervised cell is predicted
exactly, and the aggregates weight each of the five reported types by its number of puzzles. As in
PTRM, shakashaka is used in training but left out of the aggregates and our checkpoint solves $64$ of its $66$
validation puzzles and all $12$ golden ones. We evaluate the model every $1{,}000$ epochs on the
twelve evaluation sets (six types, validation and golden) and keep the last checkpoint, which is also
the best of the $21$ evaluations on the validation aggregate.

\begin{table}[h]
\caption{\textbf{Our PPBench splits against those of \citet{sghaier2026ptrm}.} Entries count puzzles per type after the size filter. PTRM's Table~4 counts training examples, which we divide by $11$. Each build draws its own validation set before the size filter, hence the different validation counts. The asterisk marks shakashaka, which is used in training but left out of the aggregates.}
\label{tab:ppb-counts}
\centering
\small
\setlength{\tabcolsep}{6pt}
\begin{tabular}{lrrrrr}
\toprule
& \multicolumn{2}{c}{training} & \multicolumn{2}{c}{validation} & golden \\
\cmidrule(lr){2-3} \cmidrule(lr){4-5} \cmidrule(l){6-6}
type & ours & PTRM & ours & PTRM & both \\
\midrule
sudoku & $708$ & $710$ & $99$ & $97$ & $15$ \\
lightup & $874$ & $864$ & $55$ & $65$ & $8$ \\
nurikabe & $1{,}446$ & $1{,}380$ & $59$ & $55$ & $9$ \\
heyawake & $4{,}086$ & $3{,}828$ & $69$ & $70$ & $7$ \\
tapa & $333$ & $333$ & $26$ & $26$ & $10$ \\
shakashaka$^\ast$ & $1{,}878$ & $1{,}882$ & $66$ & $62$ & $12$ \\
\midrule
total & $9{,}325$ & $8{,}997$ & $374$ & $375$ & $61$ \\
five reported types & & & $308$ & $313$ & $49$ \\
training examples ($\times 11$) & $102{,}575$ & $98{,}967$ & & & \\
\bottomrule
\end{tabular}
\end{table}

\begin{table}[h]
\caption{\textbf{Training configuration of the two checkpoints.} Centred entries apply to both. The loss uses the stablemax of \citet{prieto2025grokking}. The $97$ positions of TRM-MLP are the $P = 16$ prefix positions and the $81$ cells. Each checkpoint is the last of its training run.}
\label{tab:ckpt-config}
\centering
\footnotesize
\setlength{\tabcolsep}{5pt}
\fitwidth{%
\begin{tabular}{@{}lll@{}}
\toprule
& \textbf{Sudoku-Extreme} & \textbf{PPBench} \\
\midrule
\multicolumn{3}{@{}l}{\emph{model}} \\
variant & TRM-MLP (MLP across the $97$ positions) & TRM-Att (self-attention, $8$ heads) \\
positional encoding & none & RoPE \\
layers, width & \multicolumn{2}{c}{$2$, $512$; channel SwiGLU width $1{,}536$} \\
$T$, $n$, $\Nsup$ & \multicolumn{2}{c}{$3$, $6$, $16$} \\
prefix $P$ & \multicolumn{2}{c}{$16$} \\
puzzle identifiers & $1$ (shared) & $7$ ($6$ types, blank) \\
vocabulary, cells & $11$, $81$ & $439$, $100$ \\
network parameters & $5.03$M & $7.27$M \\
precision & \multicolumn{2}{c}{bfloat16 forward} \\
\midrule
\multicolumn{3}{@{}l}{\emph{training}} \\
loss & \multicolumn{2}{c}{stablemax cross-entropy $+\ 0.5 \times$ halting BCE} \\
halting exploration & \multicolumn{2}{c}{$0.1$} \\
optimiser, network & \multicolumn{2}{c}{AdamW, $\beta = (0.9, 0.95)$, decay $1.0$} \\
optimiser, puzzle embeddings & \multicolumn{2}{c}{sign-SGD, decay $1.0$} \\
learning rates & \multicolumn{2}{c}{$10^{-4}$, $2{,}000$ warm-up steps, then constant} \\
batch & \multicolumn{2}{c}{$768$} \\
training set & $1{,}000$ puzzles $\times$ $1{,}001$ examples & $9{,}325$ puzzles $\times$ $11$ examples \\
epochs & $50{,}000$ & $21{,}000$ \\
EMA & \multicolumn{2}{c}{$0.999$ on the network weights (evaluation and checkpoint)} \\
\bottomrule
\end{tabular}}
\end{table}

\FloatBarrier
\section{PTRM on Our Checkpoints}
\label{app:ptrm}
\label{app:sweep}
\label{app:gatedwidth}

\paragraph{Implementation.} Each rollout starts from the model's own initial states, so that
with $\sigma = 0$ and $D = 16$ it reproduces the standard pass bit for bit, on grids and on logits.
Noise of scale $\sigma$ is added to $\z$ before every supervision step, on all positions including the prefix, following our reading of \citet{sghaier2026ptrm}.

\paragraph{The noise scale.} We choose the PPBench scale on the validation split alone, using
best-$Q$@$25$ at the settings of PTRM's noise ablation (its Appendix~B, which does not cover
PPBench, our Table~\ref{tab:sigma-sweep}). From $\sigma = 1.25$ upwards, the five scales span $1.84$ points,
or about six instances out of $308$. The chosen $\sigma^\star = 1.5$ trails the maximum of the grid, reached at its
edge ($\sigma = 2.5$), by $0.65$ points. This is less than the $0.76$ points separating two
independent draws of the same configuration (the sweep, and the first $25$ of the $100$ rollouts of
the main rerun), so at this resolution $\sigma^\star = 1.5$ cannot be told apart from the best scale of the grid. At every scale of the grid, PTRM's pass@$25$ oracle stays below \GAR{} at $R = 16$ ($81.28\%$, Table~\ref{tab:iso}), with a maximum of $79.98\%$ at $\sigma = 2.5$.
The scale is chosen at $K = 25$, whereas the matched-retry comparison spans $K' = 1$ to $K' = 64$,
where a smaller scale might serve PTRM better: at $\sigma^\star$, a single rollout is below the
standard pass on average (Table~\ref{tab:sweep-pp}). On the golden split, PTRM is run only at the frozen scale, and this split never enters the choice of $\sigma$.

\begin{table}[h]
\caption{\textbf{The noise scale of PTRM on our PPBench checkpoint.} Pass@$25$ and best-$Q$@$25$ (\%) on the $308$ validation puzzles, with $K = 25$ rollouts of $D = 16$ steps, averaged over three seeds. The grid extends PTRM's window $[0, 1]$, beyond which this checkpoint still improves. Bold entries mark the plateau of best-$Q$@$25$ inside which we fix $\sigma^\star = 1.5$.}
\label{tab:sigma-sweep}
\centering
\small
\setlength{\tabcolsep}{8pt}
\begin{tabular}{lrr}
\toprule
$\sigma$ & pass@$25$ & best-$Q$@$25$ \\
\midrule
$0.0$ \emph{(deterministic)} & $68.83$ & $68.83$ \\
$0.1$ & $76.19$ & $76.19$ \\
$0.2$ \emph{(PTRM's value)} & $75.97$ & $75.87$ \\
$0.3$ & $76.19$ & $76.08$ \\
$0.5$ & $76.30$ & $76.19$ \\
$0.8$ & $77.16$ & $76.84$ \\
$1.0$ & $77.16$ & $76.62$ \\
\midrule
$1.25$ & $78.25$ & $\mathbf{77.92}$ \\
$1.5$ \emph{(used throughout)} & $79.00$ & $\mathbf{78.57}$ \\
$1.75$ & $78.25$ & $\mathbf{77.38}$ \\
$2.0$ & $79.87$ & $\mathbf{79.11}$ \\
$2.5$ & $79.98$ & $\mathbf{79.22}$ \\
\bottomrule
\end{tabular}
\end{table}

\paragraph{Fidelity of the rerun.} On Sudoku-Extreme, our standard pass differs slightly from PTRM's ($86.50\%$ vs.~$87.28\%$ on the full test split), and at PTRM's headline configuration the rerun gains $12.67$ points over our standard pass on our $1{,}000$ puzzles, compared with $11.47$ computed from PTRM's Table~3. On PPBench, our standard pass, which adds no noise, already differs from PTRM's in accuracy ($68.83\%$ vs.~$76.4\%$ on validation, where each paper draws its own split, and $67.35\%$ vs.~$62.6\%$ on golden), and PTRM gains less on our checkpoint than in its paper: at its headline budget, it gains $11.15$ points over our standard pass on validation and $10.20$ on golden ($49$ puzzles), versus $14.0$ and $28.6$ computed from PTRM's Tables~5 and~1. Both gaps may come from the tokenisation (our vocabulary has $439$ tokens vs.~PTRM's $294$), from the length of training, or from other differences in data or training (Appendix~\ref{app:checkpoints}), although our splits closely match PTRM's per-type counts (Table~\ref{tab:ppb-counts}). The best noise scales on our PPBench checkpoint (Table~\ref{tab:sigma-sweep}) are also several times the $0.2$ that PTRM uses there. The same implementation of the noise reproduces PTRM's headline gain on Sudoku-Extreme at PTRM's own scale, and PTRM's selected scales themselves range from $0.2$ to $1.0$ across its four benchmarks, so this difference, too, more likely reflects the checkpoint than the implementation. Within PTRM's window, the scale matters little on this checkpoint: best-$Q$@$25$ varies by less than one point between $\sigma = 0.1$ and $\sigma = 1.0$. Our comparison does not depend on the accuracies PTRM reports, since \GAR{} and our rerun use the same checkpoints, instances and evaluation path.

\paragraph{The harness applied to PTRM.} Nothing in the gate, the fallback or the stopping rule is
specific to \GAR, and withholding them from PTRM would compare systems rather than mechanisms.
Tables~\ref{tab:sweep} and~\ref{tab:sweep-pp} therefore evaluate ten PTRM configurations under each rule, on the same rollouts. For PTRM, the fallback adds the standard-pass grid to the
candidates and breaks ties in its favour, and the stopping rule goes through the rollouts of a re-opened puzzle
in the order they were drawn, stopping at the first one with $\qh \ge 0$. PTRM's own protocol (best-$Q$@$K'$) costs $K'D/16$ blocks per puzzle, and $1 + K'D/16$ with the fallback, which also runs the standard pass. Under the gate, the $K'$ rollouts of a re-opened puzzle run in parallel, for $1 + gK'D/16$ blocks, with $g$ the fraction of puzzles the gate re-opens ($0.140$ on Sudoku-Extreme and $0.273$ on PPBench validation). With the stopping rule, they run one at a time, and we measure the sequential cost puzzle by puzzle. On Sudoku-Extreme, the parallel and the sequential variants give the same answers, and on PPBench validation they differ by at most one instance.

\paragraph{Sudoku-Extreme.} The fallback removes all $29$ losses of $K' = 1$, $D = 16$, and with it at most $0.3$ losses remain in any configuration of Table~\ref{tab:sweep}. The head is thus a near-perfect
ranker here, and the $29$ losses come from the protocol, which reconsiders a puzzle whose answer was
already right. Under the gate and the stopping rule, every configuration stays within $0.10$ points of its value with the fallback. The stopping rule does make PTRM's rollouts sequential. But without it, with the
$K'$ rollouts run in parallel on the re-opened puzzles only, the configuration $K' = 10$, $D = 16$
costs more than \GAR{} at $R = 16$ and is less accurate (Tables~\ref{tab:main} and~\ref{tab:sweep}). The retries of \GAR{}
are independent draws and can run in parallel as well: keeping the first accepted retry (or the
fallback's grid if none is) returns the grids of the sequential run, at $1 + gR$ blocks. At $R = 10$, \GAR{} then costs
the same $2.40$ blocks as that configuration and reaches $97.33\%$, versus $96.37\%$
(Table~\ref{tab:iso-full}). The comparison thus does not rest on making PTRM's rollouts sequential.

\paragraph{PPBench.} At $R = 16$, \GAR{} exceeds the pass@$K'$ oracle of every configuration costing
up to $75$ blocks. Here too, the fallback removes most losses. The gate and the stopping rule keep
every configuration within three instances of its value with the fallback, and they save most of the
compute of the large ones. Without the stopping rule, the configuration $K' = 10$, $D = 16$ is again both
costlier and less accurate than \GAR{} at $R = 16$. Run in parallel, at the same $3.73$ blocks,
\GAR{} with $R = 10$ reaches $79.76\%$ (Table~\ref{tab:iso-full}), compared with $76.08\%$.

\begin{table}[H]
\caption{\textbf{PTRM on Sudoku-Extreme, with its own protocol and under the harness.} Each configuration uses the first $K'$ of $100$ rollouts of $D$ steps at $\sigma = 0.3$, with three seeds, on our $1{,}000$ test puzzles. Costs are in blocks per puzzle. Pass@$K'$ counts a puzzle as solved when any of its $K'$ rollouts is correct. \emph{Lost} counts the standard-pass successes a configuration loses, out of $859$, and is zero on every row of the bottom part, under the gate. The last columns are PTRM under the harness.}
\label{tab:sweep}
\centering
\footnotesize
\setlength{\tabcolsep}{4pt}
\begin{tabular}{lrrrrrr}
\toprule
& \multicolumn{4}{c}{own protocol} & \multicolumn{2}{c}{$+$ fallback} \\
\cmidrule(lr){2-5} \cmidrule(l){6-7}
\textbf{configuration} & cost & accuracy (\%) & pass@$K'$ & lost & accuracy (\%) & lost \\
\midrule
TRM, standard pass & $1$ & $85.90$ & --- & --- & --- & --- \\
TRM, uniform unroll ($D{=}64$) & $4$ & $92.80$ & --- & $0.0$ & --- & --- \\
\midrule
$K'{=}1$,   $D{=}16$ & $1$ & $87.70${\scriptsize$\,\pm0.44$} & $87.70${\scriptsize$\,\pm0.44$} & $29.0$ & $90.57${\scriptsize$\,\pm0.25$} & $0.0$ \\
$K'{=}5$,   $D{=}16$ & $5$ & $94.70${\scriptsize$\,\pm0.26$} & $94.70${\scriptsize$\,\pm0.26$} & $3.7$ & $95.03${\scriptsize$\,\pm0.15$} & $0.0$ \\
$K'{=}10$,  $D{=}16$ & $10$ & $96.30${\scriptsize$\,\pm0.10$} & $96.33${\scriptsize$\,\pm0.15$} & $1.3$ & $96.40${\scriptsize$\,\pm0.10$} & $0.0$ \\
$K'{=}25$,  $D{=}16$ & $25$ & $97.40${\scriptsize$\,\pm0.17$} & $97.43${\scriptsize$\,\pm0.21$} & $1.0$ & $97.43${\scriptsize$\,\pm0.21$} & $0.3$ \\
$K'{=}100$, $D{=}16$ & $100$ & $98.07${\scriptsize$\,\pm0.21$} & $98.10${\scriptsize$\,\pm0.20$} & $0.3$ & $98.07${\scriptsize$\,\pm0.21$} & $0.3$ \\
\midrule
$K'{=}1$,   $D{=}64$ & $4$ & $93.60${\scriptsize$\,\pm0.46$} & $93.60${\scriptsize$\,\pm0.46$} & $5.0$ & $94.00${\scriptsize$\,\pm0.36$} & $0.0$ \\
$K'{=}5$,   $D{=}64$ & $20$ & $96.90${\scriptsize$\,\pm0.10$} & $96.90${\scriptsize$\,\pm0.10$} & $0.3$ & $96.87${\scriptsize$\,\pm0.15$} & $0.0$ \\
$K'{=}10$,  $D{=}64$ & $40$ & $97.70${\scriptsize$\,\pm0.20$} & $97.70${\scriptsize$\,\pm0.20$} & $0.3$ & $97.67${\scriptsize$\,\pm0.12$} & $0.0$ \\
$K'{=}25$,  $D{=}64$ & $100$ & $98.07${\scriptsize$\,\pm0.12$} & $98.07${\scriptsize$\,\pm0.12$} & $0.0$ & $98.03${\scriptsize$\,\pm0.15$} & $0.0$ \\
$K'{=}100$, $D{=}64$ & $400$ & $98.57${\scriptsize$\,\pm0.06$} & $98.57${\scriptsize$\,\pm0.06$} & $0.0$ & $98.53${\scriptsize$\,\pm0.06$} & $0.0$ \\
\bottomrule
\end{tabular}

\vspace{0.7em}
\begin{tabular}{lrrrr}
\toprule
& \multicolumn{2}{c}{$+$ gate $+$ fallback} & \multicolumn{2}{c}{$+$ stopping rule} \\
\cmidrule(lr){2-3} \cmidrule(l){4-5}
\textbf{configuration} & cost & accuracy (\%) & cost & accuracy (\%) \\
\midrule
$K'{=}1$,   $D{=}16$ & $1.14$ & $90.57${\scriptsize$\,\pm0.25$} & $1.14$ & $90.57${\scriptsize$\,\pm0.25$} \\
$K'{=}5$,   $D{=}16$ & $1.70$ & $95.00${\scriptsize$\,\pm0.20$} & $1.42$ & $95.00${\scriptsize$\,\pm0.20$} \\
$K'{=}10$,  $D{=}16$ & $2.40$ & $96.37${\scriptsize$\,\pm0.15$} & $1.63$ & $96.37${\scriptsize$\,\pm0.15$} \\
$K'{=}25$,  $D{=}16$ & $4.50$ & $97.40${\scriptsize$\,\pm0.17$} & $2.06$ & $97.40${\scriptsize$\,\pm0.17$} \\
$K'{=}100$, $D{=}16$ & $15.00$ & $98.00${\scriptsize$\,\pm0.20$} & $3.58$ & $98.00${\scriptsize$\,\pm0.20$} \\
\midrule
$K'{=}1$,   $D{=}64$ & $1.56$ & $94.00${\scriptsize$\,\pm0.36$} & $1.56$ & $94.00${\scriptsize$\,\pm0.36$} \\
$K'{=}5$,   $D{=}64$ & $3.80$ & $96.83${\scriptsize$\,\pm0.12$} & $2.27$ & $96.83${\scriptsize$\,\pm0.12$} \\
$K'{=}10$,  $D{=}64$ & $6.60$ & $97.63${\scriptsize$\,\pm0.15$} & $2.79$ & $97.63${\scriptsize$\,\pm0.15$} \\
$K'{=}25$,  $D{=}64$ & $15.00$ & $97.97${\scriptsize$\,\pm0.12$} & $4.02$ & $97.97${\scriptsize$\,\pm0.12$} \\
$K'{=}100$, $D{=}64$ & $57.00$ & $98.47${\scriptsize$\,\pm0.06$} & $8.65$ & $98.47${\scriptsize$\,\pm0.06$} \\
\bottomrule
\end{tabular}
\end{table}

\filbreak
\begin{table}[H]
\caption{\textbf{The same configurations on PPBench validation.} The table follows the conventions of Table~\ref{tab:sweep} on the $308$ validation puzzles, with $\sigma^\star = 1.5$, and omits the standard deviation where the three seeds agree. \emph{Lost} counts the losses among the $212$ standard-pass successes and is zero on every row of the bottom part.}
\label{tab:sweep-pp}
\centering
\footnotesize
\setlength{\tabcolsep}{4pt}
\begin{tabular}{lrrrrrr}
\toprule
& \multicolumn{4}{c}{own protocol} & \multicolumn{2}{c}{$+$ fallback} \\
\cmidrule(lr){2-5} \cmidrule(l){6-7}
\textbf{configuration} & cost & accuracy (\%) & pass@$K'$ & lost & accuracy (\%) & lost \\
\midrule
TRM, standard pass & $1$ & $68.83$ & --- & --- & --- & --- \\
TRM, uniform unroll ($D{=}48$) & $3$ & $68.18$ & --- & $10.0$ & --- & --- \\
\midrule
$K'{=}1$,   $D{=}16$ & $1$ & $68.29${\scriptsize$\,\pm0.75$} & $68.29${\scriptsize$\,\pm0.75$} & $12.0$ & $72.19${\scriptsize$\,\pm0.75$} & $0.0$ \\
$K'{=}5$,   $D{=}16$ & $5$ & $74.13${\scriptsize$\,\pm0.82$} & $74.35${\scriptsize$\,\pm0.86$} & $3.0$ & $75.00${\scriptsize$\,\pm0.32$} & $0.3$ \\
$K'{=}10$,  $D{=}16$ & $10$ & $76.19${\scriptsize$\,\pm0.75$} & $76.62${\scriptsize$\,\pm0.56$} & $1.3$ & $76.41${\scriptsize$\,\pm0.37$} & $0.7$ \\
$K'{=}25$,  $D{=}16$ & $25$ & $77.81${\scriptsize$\,\pm0.19$} & $78.57${\scriptsize$\,\pm0.32$} & $1.0$ & $77.92${\scriptsize$\,\pm0.32$} & $0.7$ \\
$K'{=}100$, $D{=}16$ & $100$ & $80.63${\scriptsize$\,\pm0.50$} & $81.71${\scriptsize$\,\pm0.37$} & $1.3$ & $80.63${\scriptsize$\,\pm0.50$} & $1.3$ \\
\midrule
$K'{=}1$,   $D{=}48$ & $3$ & $68.72${\scriptsize$\,\pm0.75$} & $68.72${\scriptsize$\,\pm0.75$} & $15.7$ & $73.70${\scriptsize$\,\pm0.86$} & $0.3$ \\
$K'{=}5$,   $D{=}48$ & $15$ & $73.92${\scriptsize$\,\pm0.37$} & $74.57${\scriptsize$\,\pm0.37$} & $11.3$ & $77.06${\scriptsize$\,\pm0.37$} & $1.3$ \\
$K'{=}10$,  $D{=}48$ & $30$ & $75.87${\scriptsize$\,\pm0.50$} & $76.73${\scriptsize$\,\pm0.68$} & $10.3$ & $78.79${\scriptsize$\,\pm0.50$} & $1.3$ \\
$K'{=}25$,  $D{=}48$ & $75$ & $76.84${\scriptsize$\,\pm0.50$} & $78.25${\scriptsize$\,\pm0.32$} & $10.3$ & $79.65${\scriptsize$\,\pm0.37$} & $1.7$ \\
$K'{=}100$, $D{=}48$ & $300$ & $79.98${\scriptsize$\,\pm0.50$} & $82.03${\scriptsize$\,\pm0.37$} & $10.3$ & $82.47${\scriptsize$\,\pm0.86$} & $2.7$ \\
\bottomrule
\end{tabular}

\vspace{0.7em}
\begin{tabular}{lrrrr}
\toprule
& \multicolumn{2}{c}{$+$ gate $+$ fallback} & \multicolumn{2}{c}{$+$ stopping rule} \\
\cmidrule(lr){2-3} \cmidrule(l){4-5}
\textbf{configuration} & cost & accuracy (\%) & cost & accuracy (\%) \\
\midrule
$K'{=}1$,   $D{=}16$ & $1.27$ & $72.08${\scriptsize$\,\pm0.56$} & $1.27$ & $72.08${\scriptsize$\,\pm0.56$} \\
$K'{=}5$,   $D{=}16$ & $2.36$ & $74.57${\scriptsize$\,\pm0.50$} & $2.15$ & $74.57${\scriptsize$\,\pm0.50$} \\
$K'{=}10$,  $D{=}16$ & $3.73$ & $76.08${\scriptsize$\,\pm0.37$} & $3.14$ & $75.87${\scriptsize$\,\pm0.50$} \\
$K'{=}25$,  $D{=}16$ & $7.82$ & $77.27${\scriptsize$\,\pm0.32$} & $5.79$ & $77.27${\scriptsize$\,\pm0.32$} \\
$K'{=}100$, $D{=}16$ & $28.27$ & $79.98${\scriptsize$\,\pm0.50$} & $16.82$ & $79.98${\scriptsize$\,\pm0.50$} \\
\midrule
$K'{=}1$,   $D{=}48$ & $1.82$ & $73.48${\scriptsize$\,\pm1.04$} & $1.82$ & $73.48${\scriptsize$\,\pm1.04$} \\
$K'{=}5$,   $D{=}48$ & $5.09$ & $76.95$ & $4.26$ & $76.95$ \\
$K'{=}10$,  $D{=}48$ & $9.18$ & $78.14${\scriptsize$\,\pm0.82$} & $6.88$ & $77.92${\scriptsize$\,\pm0.65$} \\
$K'{=}25$,  $D{=}48$ & $21.45$ & $79.22${\scriptsize$\,\pm0.32$} & $13.94$ & $79.11${\scriptsize$\,\pm0.19$} \\
$K'{=}100$, $D{=}48$ & $82.82$ & $82.36${\scriptsize$\,\pm0.37$} & $42.25$ & $82.14${\scriptsize$\,\pm0.32$} \\
\bottomrule
\end{tabular}
\end{table}

\FloatBarrier
\section{Which State to Perturb: Full Results}
\label{app:lattice-more}

Table~\ref{tab:grid} lists the values of Figure~\ref{fig:grid} with their seed standard deviations. Its continuation row is detailed in Table~\ref{tab:lattice-full}, and its restart row gives the restarts that perturb one state. With $\y$ kept, the costs we recorded for the retries of Table~\ref{tab:lattice-full} span $1.90$ to $1.97$ blocks per puzzle on Sudoku-Extreme and $4.53$ to $4.79$ on PPBench, and with $\y$ resampled, $1.65$ to $1.69$ and $3.70$ to $3.81$. The deterministic reset of Section~\ref{sec:inside}, which returns $\y$ to $\yinit$ on the fillable cells and keeps $\z$, recovers $114.0$ failures on Sudoku-Extreme and $23.0$ on PPBench validation after $16$ continued retries.

\begin{table}[h]
\caption{\textbf{The values of Figure~\ref{fig:grid}.} In each entry, the first percentage is for Sudoku-Extreme and the second for PPBench validation. The shaded entry is \GAR{}, and the restart entry with PTRM's noise is PTRM itself, with $K' = 16$ rollouts of $D = 16$ steps under the harness.}
\label{tab:grid}
\centering
\footnotesize
\setlength{\tabcolsep}{2.8pt}
\fitwidth{
\begin{tabular}{@{}l cccc@{}}
\toprule
& \multicolumn{4}{c}{\textbf{the retry}} \\
\cmidrule(l){2-5}
& \textbf{no perturbation} & \textbf{adds PTRM's noise to} $\z$ & \textbf{resamples} $\z$ & \textbf{resamples} $\y$ \\
\midrule
\textbf{continues} (depth) & $73.8$ / $15.6$ & $70.7${\scriptsize$\,\pm2.7$} / $17.7${\scriptsize$\,\pm1.0$} & $74.2${\scriptsize$\,\pm1.5$} / $19.1${\scriptsize$\,\pm1.2$} & $88.7${\scriptsize$\,\pm1.9$} / $34.4${\scriptsize$\,\pm2.8$} \\
\midrule
\textbf{restarts} (width) & $0.0$ / $0.0$ & $78.3${\scriptsize$\,\pm1.1$} / $24.7${\scriptsize$\,\pm2.2$} & $78.5${\scriptsize$\,\pm2.5$} / $21.2${\scriptsize$\,\pm1.6$} & \cellcolor{oursrow}$87.7${\scriptsize$\,\pm1.5$} / $39.9${\scriptsize$\,\pm2.2$} \\
\bottomrule
\end{tabular}
}
\end{table}

\begin{table}[h]
\caption{\textbf{Perturbations of $\z$ crossed with the resampling of $\y$, when continuing.} Entries count the failures recovered after $16$ continued retries, by what each retry does to $\z$ (rows) and, at the block boundary, to the fillable cells of $\y$ (columns). A reset returns $\z$ to $\zinit$. \emph{All cells} means every grid cell, without the $P$ prefix positions, and \emph{all positions} includes the prefix. The noise is PTRM's, at the scales of Table~\ref{tab:main}. Bold marks the retry that resamples $\y$ alone, and a dash a retry we did not run.}
\label{tab:lattice-full}
\centering
\small
\setlength{\tabcolsep}{6pt}
\fitwidth{
\begin{tabular}{@{}l cc cc@{}}
\toprule
& \multicolumn{2}{c}{\emph{Sudoku-Extreme} ($141$ failures)} & \multicolumn{2}{c}{\emph{PPBench validation} ($96$ failures)} \\
\cmidrule(lr){2-3} \cmidrule(l){4-5}
$\z$ & $\y$ kept & $\y$ resampled & $\y$ kept & $\y$ resampled \\
\midrule
kept                     & $104.0$         & $\mathbf{125.0 \pm 2.6}$ & $15.0$         & $\mathbf{33.0 \pm 2.6}$ \\
resampled, fillable cells & $104.7 \pm 2.1$ & $123.3 \pm 2.5$          & $18.3 \pm 1.2$ & $29.7 \pm 2.9$ \\
resampled, all cells     & $102.0 \pm 3.0$ & ---                      & $17.3 \pm 0.6$ & --- \\
reset, fillable cells     & $103.0$         & $126.3 \pm 2.1$          & $15.0$         & $31.7 \pm 1.5$ \\
reset, all cells         & $101.0$         & $126.7 \pm 1.5$          & $17.0$         & $29.3 \pm 2.1$ \\
reset, all positions     & $103.0$         & $125.3 \pm 2.9$          & $19.0$         & $30.7 \pm 1.5$ \\
noise before every step  & $99.7 \pm 3.8$  & $124.3 \pm 3.2$          & $17.0 \pm 1.0$ & $31.7 \pm 3.2$ \\
\bottomrule
\end{tabular}
}
\end{table}

\paragraph{Seed by seed.} On Sudoku-Extreme, continuing without perturbation recovers $104$
failures. With $\y$ carried, the three perturbations of $\z$ (resampling the fillable cells, resampling
every grid cell, and the noise) change this count by at most $7$ for any seed, whereas resampling the fillable cells of $\y$ with $\z$ kept adds between $18$ and $23$ for every seed.

\subsection{Noise on the Reasoning State in Continued Retries}
\label{app:noise}

On Sudoku-Extreme, adding PTRM's noise to $\z$ before every step of a continued retry is inert or
slightly harmful, whether $\y$ is carried or resampled. Paired by seed, it recovers $4.3 \pm 3.8$ and $0.7 \pm 2.1$ fewer failures, which is of the order of the seed-to-seed spread (Table~\ref{tab:lattice-full}). In a
separate run, we add noise to $\z$ at $\sigma \in \{0.2, 0.5, 1.0, 1.5, 2.0\}$ in the retry that carries
$\y$, on PPBench validation with three seeds each. This recovers $13.0 \pm 2.0$, $14.3 \pm 1.2$, $19.7
\pm 1.2$, $17.3 \pm 1.5$ and $18.0 \pm 2.6$ of the $96$ failures, compared with $15.0$ without perturbation
and $33.0 \pm 2.6$ with $\y$ resampled (this run draws its own noise, hence $17.3$ at $\sigma = 1.5$ instead of $17.0 \pm 1.0$ in Table~\ref{tab:lattice-full}).
At its best scale, $\sigma = 1.0$, it recovers
$26\%$ of the gain that resampling the answer state produces, and the number of failures recovered by a single resampling of $\z$ at the
block boundary falls within the range of this grid. The noise and the resampling also differ in magnitude
(Appendix~\ref{app:protocol}), but the resampling of $\z$ on the fillable cells has the same extent
and angle as the resampling of $\y$, which removes that confound. On PPBench, perturbing the reasoning
state therefore saturates well below resampling the answer state, and the contrast of
Section~\ref{sec:inside} is not an artefact of the noise scale. In the setting the noise was designed
for, namely fresh rollouts from the initial state, its effect is measured by the restart row of
Table~\ref{tab:grid} and by the comparison with a single resampling of $\z$ (Section~\ref{sec:form}).

\FloatBarrier
\section{Matched Retries across Budgets}
\label{app:iso}

Table~\ref{tab:iso-full} extends Table~\ref{tab:iso} to seven budgets, and to the golden split at $16$ and $64$ retries, and
Table~\ref{tab:iso-terms} reports the paired differences at the same budgets. We test these differences at $r = 16$ and $r = 64$, the two budgets fixed in advance (Section~\ref{sec:protocol}), and their values at the other budgets are descriptive. The $95\%$ interval of
\GAR{} minus PTRM lies above zero at every budget except $r = 2$ on Sudoku-Extreme and $r = 1$ on
PPBench validation. At $64$
retries, the disagreements between \GAR{} and the restart that resamples $\z$ (the puzzles that only one of the two solves) favour \GAR{} for
every seed, on Sudoku-Extreme, PPBench validation and the golden split alike ($49$ of $52$ and $84$ of $90$ on the first two, summed over seeds). Compared in the same way, the continuation variant is ahead of \GAR{} for every seed on Sudoku-Extreme, and \GAR{} is ahead for every seed on PPBench validation.

\begin{table}[!htbp]
\caption{\textbf{Accuracy and cost at seven retry budgets.} The table extends Table~\ref{tab:iso} and adds the golden split at $16$ and $64$ retries. Accuracies (\%) are means over three seeds, with the standard deviation shown where the seeds disagree, and costs are in blocks per puzzle. The standard pass scores $85.90\%$, $68.83\%$ and $67.35\%$ on the three splits.}
\label{tab:iso-full}
\centering
\footnotesize
\setlength{\tabcolsep}{3.5pt}
\fitwidth{
\begin{tabular}{rrrrrrrrrrr}
\toprule
& \multicolumn{2}{c}{TRM, under the harness} & \multicolumn{2}{c}{PTRM} & \multicolumn{2}{c}{\GAR} & \multicolumn{2}{c}{$\z$ instead of $\y$} & \multicolumn{2}{c}{continue instead of restart} \\
\cmidrule(lr){2-3} \cmidrule(lr){4-5} \cmidrule(lr){6-7} \cmidrule(lr){8-9} \cmidrule(l){10-11}
$r$ & accuracy (\%) & cost & accuracy (\%) & cost & accuracy (\%) & cost & accuracy (\%) & cost & accuracy (\%) & cost \\
\midrule
\multicolumn{11}{@{}l}{\emph{Sudoku-Extreme} ($1{,}000$ puzzles, $141$ failures)} \\
$1$ & $89.70$ & $1.14$ & $90.57 \pm 0.25$ & $1.14$ & $91.43 \pm 0.32$ & $1.14$ & $91.23 \pm 0.47$ & $1.14$ & $90.87 \pm 0.15$ & $1.14$ \\
$2$ & $91.80$ & $1.24$ & $92.63 \pm 0.49$ & $1.23$ & $93.27 \pm 0.74$ & $1.22$ & $93.10 \pm 0.56$ & $1.23$ & $92.67 \pm 0.21$ & $1.23$ \\
$5$ & $93.70$ & $1.46$ & $95.00 \pm 0.20$ & $1.42$ & $95.87 \pm 0.31$ & $1.39$ & $95.30 \pm 0.46$ & $1.40$ & $95.83 \pm 0.12$ & $1.41$ \\
$10$ & $95.50$ & $1.73$ & $96.37 \pm 0.15$ & $1.63$ & $97.33 \pm 0.25$ & $1.55$ & $96.47 \pm 0.42$ & $1.61$ & $97.53 \pm 0.06$ & $1.57$ \\
$16$ & $96.30$ & $1.97$ & $96.93 \pm 0.15$ & $1.82$ & $98.27 \pm 0.21$ & $1.68$ & $96.97 \pm 0.35$ & $1.80$ & $98.40 \pm 0.26$ & $1.69$ \\
$32$ & $97.10$ & $2.50$ & $97.60$ & $2.24$ & $99.07 \pm 0.06$ & $1.85$ & $97.57 \pm 0.15$ & $2.20$ & $99.17 \pm 0.21$ & $1.86$ \\
$64$ & $97.40$ & $3.33$ & $97.87 \pm 0.06$ & $2.90$ & $99.43 \pm 0.15$ & $2.03$ & $97.90$ & $2.84$ & $99.73 \pm 0.06$ & $1.98$ \\
\midrule
\multicolumn{11}{@{}l}{\emph{PPBench validation} ($308$ puzzles, $96$ failures)} \\
$1$ & $70.78$ & $1.27$ & $72.08 \pm 0.56$ & $1.27$ & $73.05 \pm 0.32$ & $1.27$ & $70.78 \pm 0.97$ & $1.27$ & $72.73$ & $1.27$ \\
$2$ & $71.43$ & $1.53$ & $73.05 \pm 0.86$ & $1.51$ & $75.11 \pm 0.50$ & $1.49$ & $72.19 \pm 0.50$ & $1.52$ & $74.46 \pm 0.37$ & $1.49$ \\
$5$ & $72.08$ & $2.26$ & $74.57 \pm 0.50$ & $2.15$ & $77.81 \pm 0.75$ & $2.04$ & $73.70 \pm 0.32$ & $2.19$ & $76.19 \pm 0.82$ & $2.05$ \\
$10$ & $73.38$ & $3.44$ & $75.87 \pm 0.50$ & $3.14$ & $79.76 \pm 0.99$ & $2.76$ & $74.46 \pm 0.50$ & $3.23$ & $78.03 \pm 0.37$ & $2.87$ \\
$16$ & $73.70$ & $4.79$ & $76.52 \pm 0.68$ & $4.25$ & $81.28 \pm 0.68$ & $3.48$ & $75.43 \pm 0.50$ & $4.42$ & $79.55 \pm 0.86$ & $3.73$ \\
$32$ & $74.35$ & $8.32$ & $77.71 \pm 0.19$ & $6.95$ & $83.77$ & $4.98$ & $76.19 \pm 0.19$ & $7.36$ & $81.17 \pm 0.65$ & $5.73$ \\
$64$ & $75.00$ & $15.20$ & $79.00 \pm 0.19$ & $11.86$ & $85.61 \pm 0.37$ & $7.08$ & $77.16 \pm 0.19$ & $12.92$ & $83.01 \pm 0.50$ & $9.06$ \\
\midrule
\multicolumn{11}{@{}l}{\emph{PPBench golden} ($49$ puzzles, $16$ failures)} \\
$16$ & $75.51$ & $5.16$ & $72.11 \pm 1.18$ & $4.95$ & $80.27 \pm 1.18$ & $3.51$ & $74.83 \pm 1.18$ & $5.24$ & $78.23 \pm 2.36$ & $4.14$ \\
$64$ & $77.55$ & $15.82$ & $74.83 \pm 1.18$ & $14.60$ & $81.63$ & $8.07$ & $75.51 \pm 2.04$ & $16.45$ & $82.31 \pm 1.18$ & $9.78$ \\
\bottomrule
\end{tabular}
}
\end{table}

\begin{table}[!htbp]
\caption{\textbf{Paired differences at the budgets of Table~\ref{tab:iso-full}.} Each column gives the difference in accuracy points between two methods, with a $95\%$ bootstrap interval over instances. \emph{$\y$ minus $\z$} is \GAR{} minus the same restart with $\z$ resampled instead of $\y$, and \emph{once minus noise} is that restart minus PTRM, so that their sum gives \GAR{} minus PTRM up to rounding. \emph{Continue minus restart} is \GAR{}'s continuation variant minus \GAR{}. Differences are computed before rounding.}
\label{tab:iso-terms}
\centering
\footnotesize
\setlength{\tabcolsep}{4pt}
\fitwidth{
\begin{tabular}{rrrrr}
\toprule
$r$ & $\y$ minus $\z$ & once minus noise & \GAR{} minus PTRM & continue minus restart \\
\midrule
\multicolumn{5}{@{}l}{\emph{Sudoku-Extreme} ($1{,}000$ puzzles, $141$ failures)} \\
$1$ & $+0.20$ {\scriptsize $[-0.60, +1.03]$} & $+0.67$ {\scriptsize $[-0.03, +1.37]$} & $+0.87$ {\scriptsize $[+0.07, +1.67]$} & $-0.57$ {\scriptsize $[-1.23, +0.10]$} \\
$2$ & $+0.17$ {\scriptsize $[-0.60, +0.97]$} & $+0.47$ {\scriptsize $[-0.20, +1.13]$} & $+0.63$ {\scriptsize $[-0.10, +1.37]$} & $-0.60$ {\scriptsize $[-1.27, +0.07]$} \\
$5$ & $+0.57$ {\scriptsize $[-0.23, +1.37]$} & $+0.30$ {\scriptsize $[-0.20, +0.80]$} & $+0.87$ {\scriptsize $[+0.10, +1.67]$} & $-0.03$ {\scriptsize $[-0.70, +0.63]$} \\
$10$ & $+0.87$ {\scriptsize $[+0.13, +1.63]$} & $+0.10$ {\scriptsize $[-0.33, +0.53]$} & $+0.97$ {\scriptsize $[+0.30, +1.70]$} & $+0.20$ {\scriptsize $[-0.30, +0.70]$} \\
$16$ & $+1.30$ {\scriptsize $[+0.60, +2.07]$} & $+0.03$ {\scriptsize $[-0.40, +0.47]$} & $+1.33$ {\scriptsize $[+0.67, +2.10]$} & $+0.13$ {\scriptsize $[-0.30, +0.57]$} \\
$32$ & $+1.50$ {\scriptsize $[+0.80, +2.30]$} & $-0.03$ {\scriptsize $[-0.33, +0.27]$} & $+1.47$ {\scriptsize $[+0.77, +2.27]$} & $+0.10$ {\scriptsize $[-0.30, +0.50]$} \\
$64$ & $+1.53$ {\scriptsize $[+0.87, +2.27]$} & $+0.03$ {\scriptsize $[-0.23, +0.30]$} & $+1.57$ {\scriptsize $[+0.87, +2.37]$} & $+0.30$ {\scriptsize $[+0.03, +0.60]$} \\
\midrule
\multicolumn{5}{@{}l}{\emph{PPBench validation} ($308$ puzzles, $96$ failures)} \\
$1$ & $+2.27$ {\scriptsize $[+0.76, +3.90]$} & $-1.30$ {\scriptsize $[-2.60, -0.11]$} & $+0.97$ {\scriptsize $[-0.11, +2.06]$} & $-0.32$ {\scriptsize $[-1.52, +0.87]$} \\
$2$ & $+2.92$ {\scriptsize $[+1.19, +4.76]$} & $-0.87$ {\scriptsize $[-2.06, +0.22]$} & $+2.06$ {\scriptsize $[+0.43, +3.68]$} & $-0.65$ {\scriptsize $[-2.06, +0.76]$} \\
$5$ & $+4.11$ {\scriptsize $[+2.27, +6.06]$} & $-0.87$ {\scriptsize $[-2.27, +0.43]$} & $+3.25$ {\scriptsize $[+1.41, +5.30]$} & $-1.62$ {\scriptsize $[-3.35, +0.11]$} \\
$10$ & $+5.30$ {\scriptsize $[+3.14, +7.58]$} & $-1.41$ {\scriptsize $[-3.03, 0.00]$} & $+3.90$ {\scriptsize $[+1.73, +6.17]$} & $-1.73$ {\scriptsize $[-3.68, +0.22]$} \\
$16$ & $+5.84$ {\scriptsize $[+3.57, +8.33]$} & $-1.08$ {\scriptsize $[-2.27, -0.11]$} & $+4.76$ {\scriptsize $[+2.49, +7.25]$} & $-1.73$ {\scriptsize $[-3.79, +0.32]$} \\
$32$ & $+7.58$ {\scriptsize $[+4.98, +10.39]$} & $-1.52$ {\scriptsize $[-2.92, -0.32]$} & $+6.06$ {\scriptsize $[+3.35, +8.87]$} & $-2.60$ {\scriptsize $[-4.87, -0.43]$} \\
$64$ & $+8.44$ {\scriptsize $[+5.52, +11.47]$} & $-1.84$ {\scriptsize $[-3.46, -0.43]$} & $+6.60$ {\scriptsize $[+3.90, +9.52]$} & $-2.60$ {\scriptsize $[-4.76, -0.54]$} \\
\midrule
\multicolumn{5}{@{}l}{\emph{PPBench golden} ($49$ puzzles, $16$ failures)} \\
$16$ & $+5.44$ {\scriptsize $[-2.72, +14.29]$} & $+2.72$ {\scriptsize $[-0.68, +8.16]$} & $+8.16$ {\scriptsize $[+2.04, +15.65]$} & $-2.04$ {\scriptsize $[-8.84, +4.08]$} \\
$64$ & $+6.12$ {\scriptsize $[-1.36, +14.29]$} & $+0.68$ {\scriptsize $[-3.40, +6.12]$} & $+6.80$ {\scriptsize $[+0.68, +14.29]$} & $+0.68$ {\scriptsize $[-4.08, +6.12]$} \\
\bottomrule
\end{tabular}
}
\end{table}

\FloatBarrier
\section{Persistence of a Boundary Perturbation, by State}
\label{app:persistence}

The lockstep diagnostic of Table~\ref{tab:persistence} asks whether interventions on $\z$ are erased before they can act. It yields four observations. First,
resampling erases the cells it touches. In $d = 512$ dimensions, a fresh isotropic direction has
expected cosine $0$ with the vector it replaces, so, right after resampling, the mean cosine should
equal the fraction of cells left alone, which is what we find (step $0$ of Table~\ref{tab:persistence}).
Second, a perturbation of $\z$ is not too weak to act. At the end of the block, the three perturbed
copies are at about the same distance from the unperturbed copy in both states, change a
similar fraction of cells, and have each solved more re-opened puzzles than that copy, which has solved $38$. Over
$16$ retries, however, they recover $125.0$, $104.7$ and $102.0$ failures, versus $104.0$ without
perturbation (Table~\ref{tab:lattice-full}). The difference between resampling $\y$ and the two resamplings of $\z$ lies in what the later retries add. Third, the
reasoning state largely erases its own perturbation within a single step, as one would expect if the
recursion re-derives $\z$ from $\x$ and $\y$, both of which are intact: one step after all its grid cells are
resampled, $\cos \z$ is back to $0.829$. Fourth, one step after the boundary, resampling $\y$
moves $\z$ further than resampling $\z$ itself does. Across seeds, the standard deviation of the entries of Table~\ref{tab:persistence} is at most $0.026$ for the cosines, $0.025$ for the cells and $4.0$ for the counts of solved puzzles.

\begin{table}[h]
\caption{\textbf{How long a boundary perturbation survives, by state.} For each perturbation, two copies of the continuation are run in lockstep from one standard pass on the $1{,}000$ Sudoku-Extreme test puzzles, one perturbed at the block boundary and the other not. The step column counts supervision steps since the boundary, so step $16$ ends the first retry. The cosines compare the two copies cell by cell and are averaged over grid cells and re-opened puzzles, and then over seeds. \emph{Cells} gives the fraction of the fillable cells that the two copies decode differently, and \emph{solved} counts the re-opened puzzles whose decoded grid is correct at that step, out of $140$.}
\label{tab:persistence}
\centering
\footnotesize
\setlength{\tabcolsep}{3.5pt}
\fitwidth{%
\begin{tabular}{l rrrr rrrr rrrr}
\toprule
& \multicolumn{4}{c}{resampling $\y$} & \multicolumn{4}{c}{resampling $\z$, fillable cells} & \multicolumn{4}{c}{resampling $\z$, all cells} \\
\cmidrule(lr){2-5} \cmidrule(lr){6-9} \cmidrule(lr){10-13}
\textbf{step} & $\cos \y$ & $\cos \z$ & cells & solved & $\cos \y$ & $\cos \z$ & cells & solved & $\cos \y$ & $\cos \z$ & cells & solved \\
\midrule
$0$ & $0.300$ & $1.000$ & --- & $0.0$ & $1.000$ & $0.300$ & --- & $0.0$ & $1.000$ & $0.000$ & --- & $0.0$ \\
$1$ & $0.768$ & $0.565$ & $0.462$ & $1.7$ & $0.957$ & $0.837$ & $0.161$ & $4.7$ & $0.952$ & $0.829$ & $0.169$ & $5.0$ \\
$2$ & $0.892$ & $0.723$ & $0.321$ & $5.7$ & $0.929$ & $0.790$ & $0.186$ & $11.0$ & $0.934$ & $0.796$ & $0.179$ & $11.7$ \\
$4$ & $0.876$ & $0.718$ & $0.269$ & $18.7$ & $0.901$ & $0.756$ & $0.209$ & $20.0$ & $0.907$ & $0.762$ & $0.210$ & $18.0$ \\
$8$ & $0.856$ & $0.707$ & $0.257$ & $31.7$ & $0.891$ & $0.754$ & $0.219$ & $31.7$ & $0.881$ & $0.739$ & $0.228$ & $27.0$ \\
$16$ & $0.857$ & $0.720$ & $0.236$ & $49.7$ & $0.876$ & $0.747$ & $0.214$ & $48.0$ & $0.873$ & $0.736$ & $0.217$ & $43.3$ \\
\bottomrule
\end{tabular}}
\end{table}

\paragraph{Interpretation.} One step after the boundary, $\cos \z$ is back to about $0.83$ when $\z$ is resampled, and down to $0.565$ when $\y$ is resampled instead (the third and fourth observations), which suggests that, to
a first approximation, the fast state relaxes within a step towards a function of the input and the slow state. This
is reminiscent of HRM's fast module, which, according to its authors, converges within each
cycle to a local equilibrium that depends on the state of the slow module \citep{wang2025hrm}. If $\z$ were exactly such a function, the
dynamics of a continued trajectory across steps would reduce, for a given puzzle, to that of the answer state alone, and the regions that hold a failed
trajectory would be regions of answer-state space. The approximation is rough, however: by the end of the block, a resampled $\z$ has changed the grid about as much as a resampled $\y$ (the second observation). We offer this as an interpretation, subject to the limits given in Section~\ref{sec:why}.

\paragraph{The same diagnostic on PPBench.} On the $308$ validation puzzles, the first, third and fourth observations still hold. One step after the boundary, $\cos \z$ is $0.96$ when $\z$ is resampled, compared with $0.73$ when $\y$ is resampled. The second observation is weaker here. At the end of the block, the copy with $\y$ resampled is further from the unperturbed copy than the two copies with $\z$ resampled, in both states. It changes about $1.7$ times as many cells as they do, and has solved twice as many re-opened puzzles as the unperturbed copy ($12.0$ vs.~$6.0$, and $7.3$ and $6.7$ for the two resamplings of $\z$). On this benchmark, the diagnostic therefore does not say whether a perturbation of $\z$ is too weak to act. But the resampling of $\z$ on the fillable cells, which has the same extent and angle as the resampling of $\y$, and PTRM's noise up to
$\sigma = 2.0$ both recover far fewer failures than resampling $\y$ (Appendix~\ref{app:noise}).

\FloatBarrier
\section{PPBench per Type and the Golden Split}
\label{app:ppbench}

\begin{table}[H]
\caption{\textbf{PPBench per type, on the validation split.} Accuracy (\%) after $16$ and $64$ retries, averaged over three seeds. Methods are named and run as in Table~\ref{tab:iso}. Shakashaka is used in training but left out of the aggregate, as in \citet{sghaier2026ptrm}. These results are descriptive, and we run no test per type.}
\label{tab:pptype}
\centering
\footnotesize
\setlength{\tabcolsep}{3.5pt}
\fitwidth{
\begin{tabular}{lrrrrrrrrrrr}
\toprule
& & & & \multicolumn{4}{c}{$16$ retries} & \multicolumn{4}{c}{$64$ retries} \\
\cmidrule(lr){5-8} \cmidrule(l){9-12}
type & $n$ & failures & \shortstack{standard\\pass (\%)} & \GAR & \shortstack{$\z$ instead\\of $\y$} & \shortstack{continue\\instead of\\restart} & PTRM & \GAR & \shortstack{$\z$ instead\\of $\y$} & \shortstack{continue\\instead of\\restart} & PTRM \\
\midrule
sudoku & $99$ & $56$ & $43.43$ & $74.07$ & $59.93$ & $69.36$ & $60.94$ & $85.86$ & $64.98$ & $78.11$ & $67.00$ \\
lightup & $55$ & $4$ & $92.73$ & $96.97$ & $94.55$ & $96.36$ & $96.36$ & $97.58$ & $94.55$ & $97.58$ & $96.36$ \\
nurikabe & $59$ & $12$ & $79.66$ & $81.36$ & $81.36$ & $82.49$ & $82.49$ & $81.36$ & $81.36$ & $82.49$ & $83.05$ \\
heyawake & $69$ & $11$ & $84.06$ & $85.99$ & $84.06$ & $84.54$ & $85.02$ & $85.99$ & $84.06$ & $84.54$ & $86.47$ \\
tapa & $26$ & $13$ & $50.00$ & $62.82$ & $57.69$ & $62.82$ & $57.69$ & $67.95$ & $58.97$ & $67.95$ & $58.97$ \\
\cmidrule(l){1-12}
aggregate & $308$ & $96$ & $68.83$ & $81.28$ & $75.43$ & $79.55$ & $76.52$ & $85.61$ & $77.16$ & $83.01$ & $79.00$ \\
\bottomrule
\end{tabular}
}
\end{table}

\begin{table}[H]
\caption{\textbf{The main operating points on the PPBench golden split.} The split has $49$ puzzles, and the standard pass solves $33$ of them ($67.35\%$), leaving $16$ failures. The table follows the conventions of Table~\ref{tab:main}, and \emph{recovered} counts the failures that end up correct. The depth row with $\y$ resampled is the continuation variant of \GAR{}, and the width row with $\z$ resampled is ``$\z$ instead of $\y$''.}
\label{tab:golden}
\centering
\small
\setlength{\tabcolsep}{6pt}
\fitwidth{
\begin{tabular}{@{}cllrrrr@{}}
\toprule
\textbf{state} & \textbf{method} & \textbf{runs on} & \textbf{cost} & \textbf{accuracy (\%)} & \textbf{recovered} & \textbf{lost} \\
\midrule
--- & TRM, standard pass      & all      & $1$    & $67.35$          & ---   & ---   \\
\midrule
\multicolumn{7}{@{}l}{\emph{Depth: one trajectory, continued}} \\
--- & TRM, uniform unroll ($D{=}48$)   & all      & $3$    & $71.43$          & $3.0$ & $1.0$ \\
--- & no perturbation ($R{=}16$)          & rejected & $5.16$ & $75.51$          & $4.0$ & $0.0$ \\
$\y$ & resampling ($R{=}16$)            & rejected & $4.14$ & $78.23 \pm 2.36$ & $6.3$ & $1.0$ \\
\midrule
\multicolumn{7}{@{}l}{\emph{Width: trajectories drawn from the initial state}} \\
--- & no perturbation (control, $R{=}1$) & rejected & $1.33$ & $67.35$          & $0.0$ & $0.0$ \\
$\z$ & resampling ($R{=}16$)           & rejected & $5.24$ & $74.83 \pm 1.18$ & $3.7$ & $0.0$ \\
\rowcolor{oursrow} $\y$ & \textbf{\GAR{} ($R{=}16$)}       & rejected & $3.51$ & $80.27 \pm 1.18$ & $7.3$ & $1.0$ \\
\rowcolor{oursrow} $\y$ & \textbf{\GAR{} ($R{=}64$)}       & rejected & $8.07$ & $81.63$ & $8.0$ & $1.0$ \\
\midrule
\multicolumn{7}{@{}l}{\emph{PTRM with its own protocol: $K'$ rollouts (width) of $D$ steps (depth)}} \\
$\z$ & PTRM ($K'{=}1$, $D{=}16$)        & all      & $1$    & $63.95 \pm 1.18$ & $1.3$ & $3.0$ \\
$\z$ & PTRM ($K'{=}25$, $D{=}16$)       & all      & $25$   & $73.47 \pm 2.04$ & $4.0$ & $1.0$ \\
$\z$ & PTRM ($K'{=}100$, $D{=}16$)      & all      & $100$  & $76.19 \pm 1.18$ & $5.3$ & $1.0$ \\
$\z$ & PTRM ($K'{=}100$, $D{=}48$)      & all      & $300$  & $77.55$          & $7.0$ & $2.0$ \\
\bottomrule
\end{tabular}
}
\end{table}

\begin{table}[H]
\caption{\textbf{PPBench per type, on the golden split.} Accuracy (\%), averaged over three seeds, with PTRM run as in Table~\ref{tab:iso}. Lightup and nurikabe have no standard-pass failures, and both methods lose the nurikabe puzzle re-opened by the gate's only false positive. These results are descriptive.}
\label{tab:golden-type}
\centering
\footnotesize
\setlength{\tabcolsep}{6pt}
\fitwidth{
\begin{tabular}{@{}llrrrrrr@{}}
\toprule
retries & method & sudoku & lightup & nurikabe & heyawake & tapa & aggregate \\
\midrule
\rowcolor{oursrow}$16$ & \textbf{\GAR} & $71.11$ & $100.00$ & $88.89$ & $80.95$ & $70.00$ & $80.27$ \\
 & PTRM & $55.56$ & $100.00$ & $88.89$ & $71.43$ & $60.00$ & $72.11$ \\
\midrule
\rowcolor{oursrow}$64$ & \textbf{\GAR} & $75.56$ & $100.00$ & $88.89$ & $80.95$ & $70.00$ & $81.63$ \\
 & PTRM & $64.44$ & $100.00$ & $88.89$ & $71.43$ & $60.00$ & $74.83$ \\
\bottomrule
\end{tabular}
}
\end{table}

\paragraph{What limits recovery on PPBench.} The gains are concentrated where the standard pass has the
most headroom, on sudoku and tapa, and are smallest on nurikabe and heyawake, where \GAR{} recovers
$1$ of $12$ and $1.3$ of $11$ failures at both budgets (Table~\ref{tab:pptype}). None of the methods of Table~\ref{tab:pptype} loses a solved puzzle, so, for a type with $n$ puzzles, the number of failures a method recovers is $n/100$ times its gain in accuracy over the standard pass. The lower recovery on PPBench comes with four measured
differences from Sudoku-Extreme. Depth alone does much less
(Figure~\ref{fig:grid}). The budget becomes limiting, as the fraction of re-opened puzzles still active
falls far more slowly (Figure~\ref{fig:attempts}). Recall is lower (Table~\ref{tab:gate}) and, on validation, uneven
across types: $100\%$ on lightup, $98.2\%$ on sudoku, $72.7\%$ on heyawake, $69.2\%$ on tapa and
$66.7\%$ on nurikabe, with no false positive on any type. Finally, the head accepts wrong grids on
retries. At $R = 64$, at least $15$ of the $84$ re-opened validation puzzles stop on a wrong grid,
since $51.7$ of them end up correct and $17.3$ are still active before the last retry
(Table~\ref{tab:iso-full}, Section~\ref{sec:restart}). On Sudoku-Extreme, fewer than $5$ of the $140$ do, since $135.3$ of them end up correct. Most of the gap between resampling $\y$ and resampling
$\z$ comes from the sudoku type, where the restart that resamples $\z$ stays $20.9$ points below
\GAR{} at $64$ retries ($20.7$ of the $26.0$ puzzles that separate them), and from tapa. The restart that resamples $\z$ is not ahead of \GAR{} on any type, at either budget. The lead of \GAR{} over PTRM is concentrated in the same way: at $64$ retries, the sudoku type accounts for $18.7$ of the $20.3$ puzzles that separate them, and PTRM is ahead only on nurikabe and heyawake, by one puzzle and by a third of one. Even so, resampling the answer state more than doubles what
continuing without perturbation recovers at $16$ retries, a larger relative gain than on
Sudoku-Extreme (Table~\ref{tab:lattice-full}).

\paragraph{The golden split.} Table~\ref{tab:golden} gives the main operating points on the golden
split, and Tables~\ref{tab:iso-full} and~\ref{tab:iso-terms} the matched retries. \GAR{} outperforms
PTRM at $16$ and $64$ retries, with intervals that exclude zero, though at $64$ retries the lower bound is a third of a puzzle (Table~\ref{tab:iso-terms}). Per type (Table~\ref{tab:golden-type}), \GAR{} is at least as accurate as PTRM at both budgets and ahead on the three types that have failures. With only $n = 49$ puzzles, however, neither the comparison of $\y$ with $\z$ nor that of continuing with restarting is conclusive, although the point
estimates of the first comparison favour resampling $\y$ at both budgets. This split played no part in the choice of retry (Section~\ref{sec:gar}).

\end{document}